\documentclass[10pt,twocolumn,letterpaper]{article}

\usepackage[pagenumbers]{cvpr} 
\usepackage{amsmath}
\usepackage{subcaption}
\usepackage{multicol}
\usepackage{graphicx}
\usepackage{xcolor}
\usepackage{algpseudocode}
\usepackage{algorithm}
\usepackage{amsmath}
\usepackage{bm}
\usepackage{subfiles}
\graphicspath{{figures/}}

\usepackage[switch]{lineno}

\usepackage{tikz}
\usetikzlibrary{bayesnet}

\definecolor{cvprblue}{rgb}{0.21,0.49,0.74}
\usepackage[pagebackref,breaklinks,colorlinks,citecolor=cvprblue]{hyperref}

\newcommand{\card}[1]{\lvert\mathcal{#1}\rvert}

\def\paperID{15137} 
\def\confName{CVPR}
\def\confYear{2025}

\title{Gradient-Guided Annealing for Domain Generalization}

\author{Aristotelis Ballas\\
Dpt of Informatics and Telematics\\
Harokopio University of Athens\\
Omirou 9, Tavros, Athens, Greece\\
{\tt\small aballas@hua.gr}
\and
Christos Diou\\
Dpt of Informatics and Telematics\\
Harokopio Univesity of Athens\\
Omirou 9, Tavros, Athens, Greece\\
{\tt\small cdiou@hua.gr}
}

\begin{document}

\twocolumn[{%
	\renewcommand\twocolumn[1][]{#1}%
	\maketitle
	\begin{center}
		\setcounter{figure}{0}
		\centering
		\captionsetup{type=figure}
		
		\vspace{-1.5em}
		\includegraphics[width=0.7\textwidth,trim={0.0cm 0.0cm 0cm 0},clip,page=1]{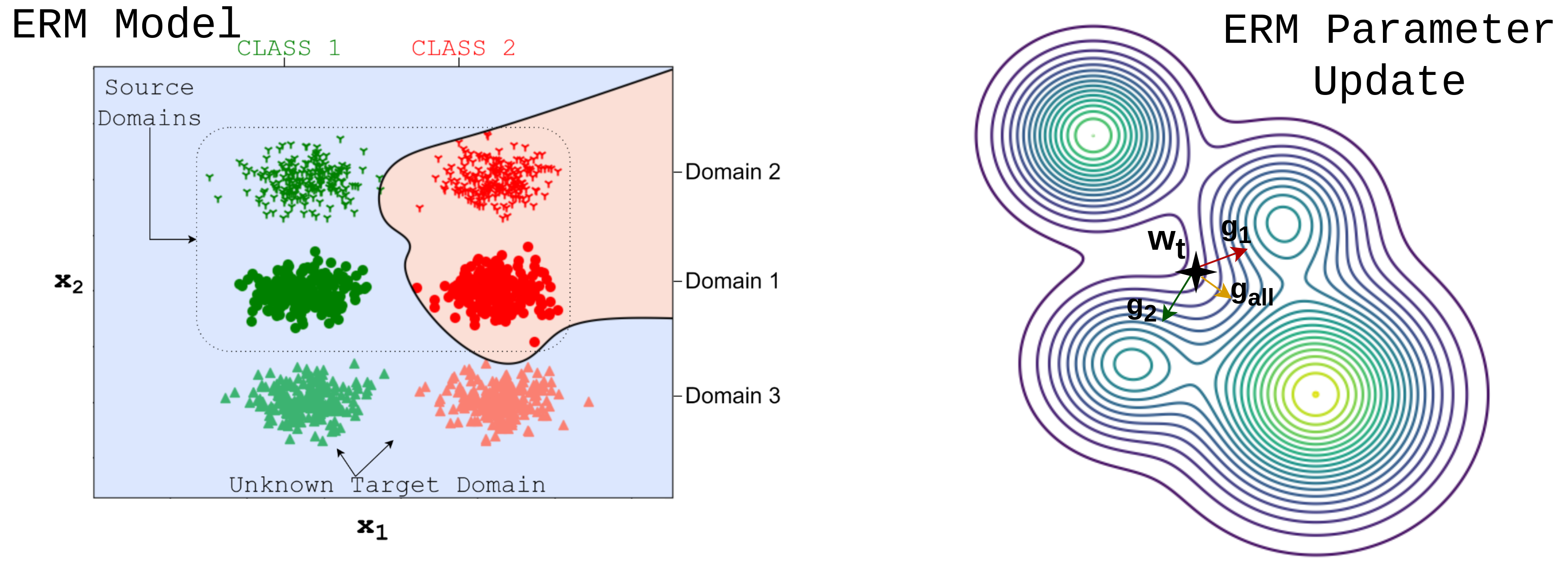}\vspace{-0.5em}%
		
		\vspace{1em}
		
		\includegraphics[width=0.7\textwidth,trim={0.0cm 0.0cm 0cm 0},clip,page=1]{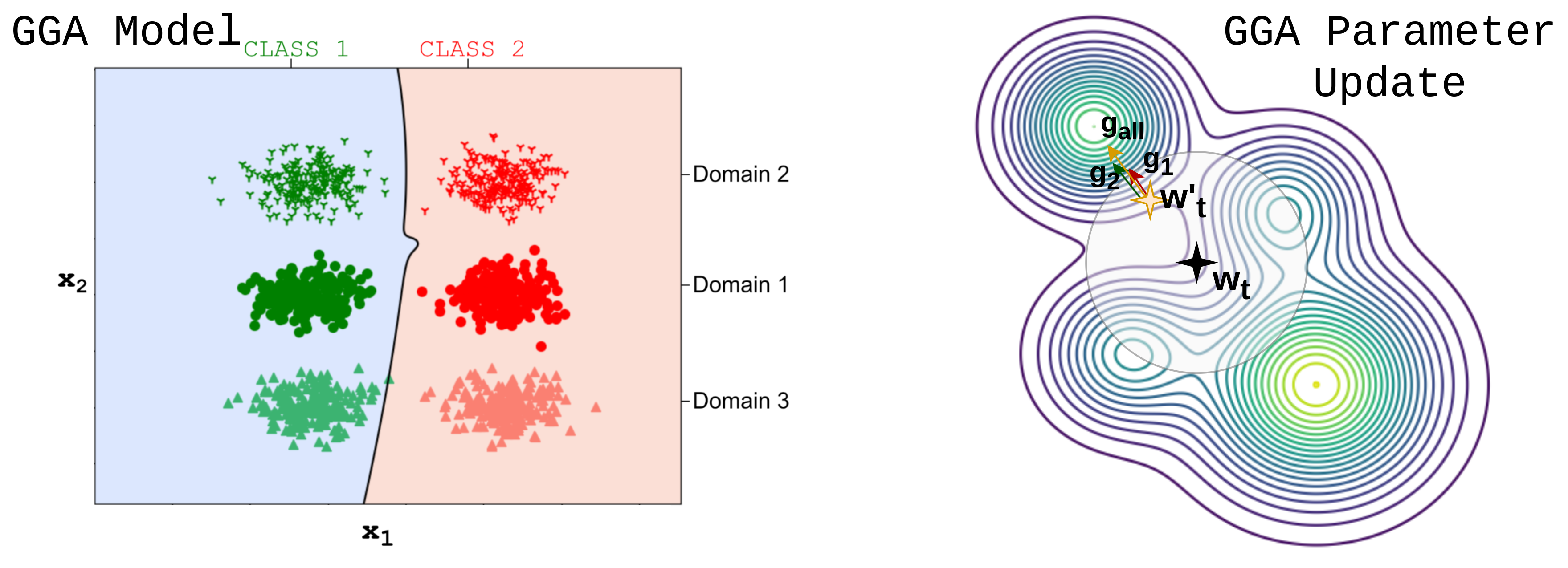}\vspace{-0.5em}%
		\caption{\em (left) Decision boundaries of a 4\textsuperscript{th}-degree polynomial logistic regression model with 2D input. In this example, feature $x_1$ is class-specific and $x_2$ is domain-specific, while color represents classes and shapes represent domains. The samples with solid red and green colors are included in the training data, whereas the fainted samples are part of the hidden held-out test set. As a result, domain shift is represented by a change in $x_2$. Although the classifier should only infer based on $x_1$, traditional gradient descent leads to overfitting (top-left). The proposed method, \textit{GGA} (bottom-left), introduces an annealing process that depends on gradient agreement, leading to models that generalize well to new, unobserved target domains. (right) Schematics of the parameter updates of ERM (top-right) and GGA (bottom-right). Parameters updated via ERM are driven by gradient conflict, whereas GGA searches for a point where gradients align before continuing descending towards a minima.}\label{fig:concept}%
	\end{center}%
}]

\begin{abstract}

Domain Generalization (DG) research has gained considerable traction as of 
late, since the ability to generalize to unseen data distributions is a 
requirement that eludes even state-of-the-art training algorithms. In this 
paper we observe that the initial iterations of model training play a key 
role in domain generalization effectiveness, since the loss landscape may be 
significantly different across the training and test distributions, contrary 
to the case of i.i.d. data. Conflicts between gradients of the loss 
components of each domain lead the optimization procedure to undesirable 
local minima that do not capture the domain-invariant features of the target 
classes. We propose alleviating domain conflicts in model optimization, by 
iteratively annealing the parameters of a model in the early stages of 
training and searching for points where gradients align between domains. By 
discovering a set of parameter values where gradients are updated towards the 
same direction for each data distribution present in the training set, the 
proposed Gradient-Guided Annealing (GGA) algorithm encourages models to seek 
out minima that exhibit improved robustness against domain shifts. The 
efficacy of GGA is evaluated on five widely accepted and challenging
image classification domain generalization benchmarks, where its use alone is
able to establish highly competitive performance against the state-of-the-art.
Moreover, when combined with previously proposed domain-generalization
algorithms it is able to consistently improve their effectiveness by significant margins\footnote{Code available at: \href{https://github.com/aristotelisballas/GGA}{https://github.com/aristotelisballas/GGA}}.

\end{abstract}

\section{Introduction}
\label{sec:intro}

The vast majority of neural networks today are trained via stochastic gradient
descent methods, such as SGD \cite{bottou1998online} or ADAM
\cite{kingma2014adam}, where the gradient direction guides the optimization
through the loss landscape, aiming to converge to a minimum. What is more, it
has been empirically shown that the astounding performance and generalization
capabilities of these over-parameterized models stems from the fact that loss
surfaces of large neural networks have multitudinous local minima
\cite{auer1995exponentially, safran2016quality}, most of which yield similar
performance upon model convergence \cite{choromanska2015loss,
  nguyen2018optimization}.

The above optimization process assumes that all available data samples are independent and identically
distributed (i.i.d.). When this assumption holds, the parameter values reached
during training is likely to generalize to the test distribution. However, in
several real-world scenarios the i.i.d. assumption is not met, and models are
evaluated on similar but distinct out-of-distribution (OOD) data resulting 
from domain shifts to the training distribution, leading to a domain
generalization (DG) \cite{zhou_domain_2023} problem. In this case, the 
training loss minimum may lead to a poor parameter configuration for the test data.

The disagreement of gradients across data from different domains during 
training, can provide an indication that the described problem occurs. This observation 
about gradient conflicts was initially made in the multi-task learning 
paradigm \cite{9392366}, where gradients of different tasks pointed to 
conflicting directions on the loss surface. To mitigate the effects of 
gradient conflicts, methods to balance the relative gradient magnitudes were 
proposed \cite{chen2018gradnorm, kendall2018multi}, along with algorithms that remove disagreeing gradient components among tasks \cite{sener2018multi,
  yu2020gradient}. Similarly, gradient conflicts were also addressed in the
context of DG \cite{mansilla2021domain}, where gradients with different signs among domains were either muted or set to random values; as they are 
assumed to contain domain-specific information. With that being said, it has 
been empirically shown that even though the above issues arise, the correct 
selection and tuning of hyperparameters can be enough for vanilla network 
training to surpass even state-of-the-art algorithms \cite{gulrajani2020domainbed}. This evidence leads us to believe that there 
exist local minima on the vanilla loss surface that lead to more robust and 
generalizable models. Depending on the starting parameter configuration, even a 
model optimized by traditional Empirical Risk Minimization can reach a 
solution that is locally optimal for all distinct domains. These hypotheses, 
along with problems poised by gradient disagreement, are demonstrated in Section \ref{toy-example} (illustrated in Fig.\ref{fig:concept}) and are further explored in Section \ref{sec:generative}.

Starting from these ideas, we propose an alternative strategy for updating the
parameters of a neural network during the optimization process, in an attempt to
``\textit{set the model up for success}''. Specifically, inspired by
Simulated Annealing Optimization \cite{kirkpatrick1983optimization}, during the
early stages of model training we iteratively anneal, i.e. randomly perturb the
parameters of the model, and search for a set of parameter values where the
gradients between all training domains agree, before minimizing the total loss
across all training domains. We call this simple, yet effective, strategy
\textit{Gradient-Guided Annealing} or \textit{GGA}. When evaluated on extensive
and challenging DG benchmarks, GGA is able to boost the performance of the
baseline by significant margins, even yielding results competitive to the state-of-the-art without the application of additional data processing or 
augmentation techniques. Furthermore, since GGA can be considered a general strategy for handling multi-domain datasets it can also be combined with previously proposed algorithms. Experimental results of the combined methods,
demonstrate the efficacy of GGA, as it is able to enhance their
generalization capacity and overall accuracy, boosting their performance over the baseline benchmarks.

Our primary contributions are as follows:
\begin{itemize}
 \item We present Gradient-Guided Annealing (GGA), a DG method for training neural networks such that gradients align across domains.
 \item We validate the effectiveness of GGA on multiple, challenging Domain Generalization benchmarks both as a standalone algorithm and by combining it with previous state-of-the-art methods.
 \item We offer further evidence on the effectiveness of the proposed method by investigating the domain gradients during training and its sensitivity to the choice of hyperparameters.
\end{itemize}


\subsection{Gradient Disagreement and Domain Overfitting}
\label{toy-example}

To demonstrate how the domain generalization problem manifests during
optimization in the source domains during training, we consider the following
simple synthetic binary classification problem. For training, data is sampled
from a mixture of two domains, while testing takes place on a different,
previously unseen, target domain. Each sample has two features, a class-specific
feature $x_1$ and a domain-specific feature $x_2$. Our goal is to learn a model
on the source domains, that can discriminate between classes effectively on the
unseen target domain.

Concretely, for the source domains, we draw 200 points from a 2-D Gaussian
distribution with an isotropic covariance matrix for each domain and class, i.e.,
\begin{equation}
	\bm{x}_{y}^{(d)} \sim \mathcal{N}(\bm{\mu}_{y}^{(d)}, \sigma \bm{I}_2)
\end{equation}
The subscript $y$ indicates the class and the exponent $d$ the domain,
while $\bm{\mu}_1^{(1)} = [-2.5, -2.5]$, $\bm{\mu}_1^{(2)} = [-2.5, 2.5]$, $\bm{\mu}_2^{(1)} =
[2.5, -2.5]$, $\bm{\mu}_2^{(2)} = [2.5, 2.5]$, $\sigma = 0.5$ and $\bm{I}_2$ is
the $2\times 2$ identity matrix. We also draw additional 400 samples from a
held-out test domain with means at $\bm{\mu}_1^{(3)} = [-2.5, -7.5]$ and $\bm{\mu}_2^{(3)}
= [2.5, -7.5]$. The drawn samples are shown in the left column of Fig. \ref{fig:concept}. In this
example, the classes can be distinguished solely on $x_1$ (the
``class-specific'' feature), while domains differ in terms of feature $x_2$ (the
``domain-specific'' feature).

We train a 4\textsuperscript{th} degree polynomial logistic regression model
trained with Empirical Risk Minimization (ERM, \cite{vapnik1998statistical}) and
binary cross-entropy loss using the SGD optimizer. In the top-left example of Fig. \ref{fig:concept} the initial parameter conditions were such that 
training converged to a local minimum of the loss that leads the model to consider both $x_1$ and $x_2$ for its decisions. In this case, the model has clearly overfit
its source domains and will fail when presented with out-of-distribution data
from the held-out domain. An indicator of this was the fact that gradients of
the loss for samples of different domains were dissimilar during training. This
is in contrast to the bottom-left GGA model in Fig. \ref{fig:concept} which 
mostly relies on $x_1$ to discriminate between classes. This model was trained by using the
proposed gradient agreement strategy and although the resulting training loss is
slightly higher, the model successfully generalizes to new domains.

In the rest of the paper, we first discuss the most relevant works in the domain generalization literature (Section \ref{sec:related}) and then present 
the proposed methodology (Section \ref{sec:methods}). Followingly, we present the experimental setup and results (Section \ref{sec:experiments}), and finally conclude the paper with a
discussion on limitations of our method and directions for future research
(Section \ref{sec:conclusions}).

\section{Related Work}
\label{sec:related}

\textbf{Domain Generalization (DG)} methods focus on learning a model from one
or multiple \textit{source} data sets, or domains, which can generalize to
previously unseen, out-of-distribution \textit{target} domains. Existing DG
methods in the literature can be categorized into two major groups; single-source and multi-source. In addition to not having any knowledge about the unseen data,
single-source algorithms do not leverage information regarding the presence of
disinct domains in the training set. On the other hand, multi-source methods utilize
domain labels and often take advantage of the statistical differences in the
sample distributions. Specifically, most popular algorithms include data
augmentation \cite{zhou2021mixstyle, carlucci2019domain} which proves beneficial
for regularizing over-parameterized neural networks and improving
generalization, meta-learning \cite{li2018learning, zhang2020arm,
  balaji2018metareg, dou2019domain}, which exposes models to domain shifts
during training, and disentangled representation learning \cite{peng2019domain, wang2020cross, zhang2022towards, 10472869}, where 
models most commonly include modules or architectures that focus on decomposing learned representations into
domain-specific and domain-invariant parts. Additionally, domain alignment
\cite{muandet2013domain, ganin2016domain, wang2021respecting} and causal
representation learning algorithms \cite{mahajan2021domain, lv2022causality}
have also been proposed in the literature towards producing robust models that
retain their generalization capabilities on unseen data. Finally, ensemble
learning methods \cite{zhou2012ensemble} have also been proposed for DG, where
techniques such as weight averaging \cite{izmailov2018averaging} lead to
improved generalization \cite{cha2021swad}.

\textbf{Gradient operations for DG.} Lately, there has been a surging interest
in addressing the DG problem from a gradient-aware perspective. The most relevant works to ours, leverage gradient information to learn generalized
representations from the source datasets. For example, \cite{huang2020rsc}
proposes a self-challenging learning scheme, by muting the feature
representations associated with the highest gradients and forcing the network to
learn via alternative routes. In another work, the authors of
\cite{shi2021gradient} propose Fish, a first-order algorithm that approximates
the optimization of the gradient inner product between domains. Inspired by gradient surgery in multi-task learning \cite{yu2020gradient},
\cite{mansilla2021domain} proposes aligning the gradient vectors among source
domains by retaining the individual same-sign gradients, and either setting the rest to zero or random values. Finally, there has also been great interest into researching the properties of Sharpness-Aware Minimization (SAM)
\cite{foret2020sharpness, zhuang2022surrogate, wang2023sharpness,
  le2024gradient}, as it has been hypothesized that flatter minima lead to
smaller DG gaps and improved generalization.

\textbf{Simulated Annealing for Deep Learning.} Although explored in past 
literature, simulated annealing (SA) has not been explicitly proposed for DG. 
To avoid being ``trapped'' in local minima during model optimization, 
\cite{cai2021sa} proposes SA-GD, a simulated annealing method for gradient 
descent. Similarly, \cite{RERE2015137} shows that by sacrificing computation time, simulated annealing optimization can also improve the results of
standard CNN architectures. In a more recent work, the authors of \cite{sarfi2023simulated} propose SEAL and apply simulated annealing in the 
early layers of networks to prohibit them from learning overly specific 
representations and improve model generalization. 
Furthermore, there has also been research regarding the combination of SA 
with reinforcement learning (RL) algorithms, where RL is used to optimize 
specific hyperparameters of the SA process \cite{pmlr-v206-correia23a}.

When compared to previously proposed methods which take into consideration
gradient behaviour, GGA has some key differences. In contrast to gradient
surgery algorithms \cite{yu2020gradient, mansilla2021domain} that either mute,
aggregate or set gradients to random values, and Fish \cite{shi2021gradient}
that approximately optimizes the inner-product between domain gradients thus altering the loss function, GGA searches for existing parameter space points where gradients of different domains have pairwise small angles on the ERM loss surface. Furthermore, as GGA is applied in the early stages of training and for a limited number of training iterations, the high computational burden of traditional simulated annealing methods is avoided.
\section{Methods}
\label{sec:methods}

\subsection{Preliminaries}
\label{sec:preliminaries}
Consider a classification problem with $K$ classes. During model training we
have access to a data set composed of distinct source data distributions (or
\textit{domains}), $\mathcal{S} = \left\{D_1, D_2,\dots,
D_{\card{S}}\right\}$. From each domain $D_i$, we observe $n_i$
labeled data points, such that $(\bm{x}_j^{(i)}, y_j^{(i)}) \sim D_i$, for
$j=1,\dotsc,n_i$. Similarly, the test dataset consists of $\mathcal{T} =
\left\{D^{T}_1, D^{T}_2, \dotsc, D^{T}_{\card{T}}\right\}$
\textit{unseen} target data distributions, from which the model cannot retrieve
any information during training. The goal is to learn a single labeling function
$h(\bm{x}; \bm{\theta})$, parameterized by $\bm{\theta}$, which correctly maps
input observations $\bm{x}_j^{(i)}$ to their labels ${y_j^{(i)}}$ for both the
seen source and the unseen target domains.

If $\mathcal{L}_i(\bm{\theta}) = \frac{1}{n_i} \sum_{j=1}^{n_i}
\ell(h({\bm{x}_j}^{(i)};\bm{\theta}), {y_j^{(i)}})$ represents the loss associated to the
$i$-th source data domain in the training set, we define the overall cost
function $\mathcal{L}(\bm{\theta}) = \frac{1}{\card{S}} \sum_{i=1}^{\card{S}}
\mathcal{L}_i(\bm{\theta})$, as the average loss over all available source domains.
The function $\ell(\cdot, \cdot)$ is a classification loss, in our
case cross-entropy, that measures the error between the predicted label
$\hat{y}$ of an input observation and its true label $y$. The standard way of model training is \emph{Empirical Risk Minimization} or \emph{ERM} which uses the following objective on the training domain data
\begin{equation}\label{erm-optimization-eq}
	\bm{\theta}^{*} = \arg_{\bm{\theta}}\min \frac{1}{\card{S}} \sum_{i=1}^{\card{S}} \mathcal{L}_i(\bm{\theta}) + \lambda R(\bm{\theta})
\end{equation}
where $R(\cdot)$ is a regularization term and $\lambda$ a hyperparameter responsible for controlling the contribution of regularization to the loss, leading to a parameter vector $\bm{\theta}^{*}$. In practice, networks trained via ERM have been shown to overfit the data distributions present in the training set. Previous works (such as \cite{shi2021gradient}) have observed that the directions of gradients for different domains during training play a significant role in model generalization. Given source domain losses $\mathcal{L}_i (\bm{\theta})$ and $\mathcal{L}_j(\bm{\theta})$ and their corresponding gradients $\bm{g}_i = \nabla_{\bm{\theta}} \mathcal{L}_i(\bm{\theta})$, $\bm{g}_j = \nabla_{\bm{\theta}} \mathcal{L}_j(\bm{\theta})$, gradient \textbf{conflicts} arise when the angle of the gradients grows e.g., above $\lvert \pi/2 \rvert$, or, equivalently, their cosine similarity $\frac{\bm{g}^{T}_i \cdot \bm{g}_j}{\|\bm{g}_i\| \|\bm{g}_j\|}$ becomes negative. In the following section, we further explore the role of gradient similarity as an indicator for domain overfitting.


\subsection{A Generative Model for Domain Generalization}

\label{sec:generative}
\begin{figure}
  \centering
    \begin{tikzpicture}

  \node[obs]                                (X) {$\mathbf{x}$} ;
  \node[det, above=of X, yshift=-0.5cm]                                (f) {$f$} ;
  \node[latent, left=of f]                             (e) {$\mathbf{e}$} ;
  \node[latent, above=of f, xshift=-1.2cm, yshift=-0.5cm]  (Zy) {$\mathbf{z}_y$} ;
  \node[latent, above=of f, xshift=1.2cm, yshift=-0.5cm]  (Zd) {$\mathbf{z}_d$} ;
  \node[latent, above=of Zy, yshift=-0.5cm]  (Y) {$y$} ;
  \node[latent, above=of Zd, yshift=-0.5cm]  (D) {$d$} ;

  \edge{Y}{Zy}
  \edge{D}{Zd}
  \edge{Zy,Zd,e}{f}
  \edge{f}{X}

\end{tikzpicture}
    %
    \caption{Simplified generative model for multi-domain data.}
    \label{fig:generative}
\end{figure}
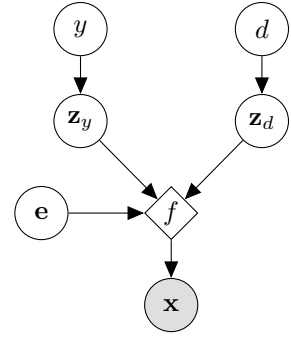

To assist with the development and understanding of the proposed method consider the generative model of Figure \ref{fig:generative}, summarized by the following process:
\begin{align*}
  y &\sim Multinoulli(p_1, \dotsc, p_K)\\
  d &\sim Multinoulli(q_1, \dotsc, q_L)\\
  \bm{z}_y &\sim p_y \\
  \bm{z}_d &\sim p_d \\
  \bm{e} &\sim p_e \\
  \bm{x} &= f(\bm{z}_y, \bm{z}_d, \bm{e})
\end{align*}
where $y$ is a variable corresponding to the class, $d$ is a variable corresponding to the domain (with $L=\card{S}+\card{T}$), $\bm{z}_y$ is a latent multivariate class-specific representation of class $y$, drawn from an unknown distribution $p_y$, and $\bm{z}_d$ is a latent multivariate domain-specific representation of the domain $d$, drawn from an unknown distribution $p_d$. Notice that these representations are disentangled, i.e., $\bm{z}_y$ does not depend on $d$. The observed sample $\bm{x}$, on the other hand, is derived using the function $f(\bm{z}_y, \bm{z}_d, \bm{e})$ that depends on both the class and domain representations, as well as independent nuisance variables, $\bm{e}$.

Without loss of generality, let's assume that we have two source domains ($\card{S} = 2$) and two classes ($K = 2$). Then, during model training we sample only from the source domains and use the loss $\mathcal{L} = \alpha_1\mathcal{L}_1 + \alpha_2\mathcal{L}_2$,
\begin{equation}
  \mathcal{L}_i = -\mathbb{E}_{d=i}\left[\log h(y | \bm{x};\bm{\theta})\right]
  \label{eq:loss}
\end{equation}
where $h$ is the probability estimated by our model for class $y$,  $\alpha_i$ is the percentage of samples of domain $i$, and $i = 1, 2$. The expectation is over samples that have been generated using the above process when $d=i$. Each gradient update step depends on the loss gradient
\begin{equation}
  \begin{split}
  \nabla_{\bm{\theta}}\mathcal{L} &= \alpha_1\nabla_{\bm{\theta}}\mathcal{L}_1 + \alpha_2\nabla_{\bm{\theta}}\mathcal{L}_2\\
  &= -\alpha_1\mathbb{E}_{d=1}\left[\nabla_{\bm{\theta}}\log h(y | \bm{x};\bm{\theta})\right] \\
  & -\alpha_2\mathbb{E}_{d=2}\left[\nabla_{\bm{\theta}}\log h(y | \bm{x};\bm{\theta})\right]
  \end{split}
  \label{eq:expgrad}
\end{equation}
where we have used the linearity of the expectation to obtain the expectation of
the derivatives. Notice that the gradient update step depends on
$\nabla_{\bm{\theta}}\log h(y | \bm{x};\bm{\theta}) = \nabla_{\bm{\theta}}\log h(y | f(\bm{z}_y, \bm{z}_d, \bm{e} ;\bm{\theta}))$.

For models that use an internal domain-invariant representation
(i.e., a representation that does not depend on $\bm{z}_d$) it will hold that
$h(y|\bm{x}) = h(y|f(\bm{z}_y, \bm{z}_1, \bm{e})) = h(y|f(\bm{z}_y,\bm{z}_2,
\bm{e}))$, and therefore $\mathbb{E}_{d=1}\left[u(h(y|\bm{x}))\right] =
\mathbb{E}_{d=2}\left[u(h(y|\bm{x}))\right]$ for any function $u$. Thus, both
the losses, $\mathcal{L}_i$ and their gradients,
$\nabla_{\bm{\theta}}\mathcal{L}_i$, should be equal, in expectation, for
different domains $i$. This observation provides a \emph{necessary} condition for
domain-invariance, which in turn provisions the main motivation for the development of our method, presented in the following section.

\subsection{Gradient-Guided Annealing for DG}
\label{sec:annealing}

\begin{algorithm}[t!]
	\caption{Implementing GGA in training}
	\begin{algorithmic}[1]
		\Require   Pretrained DNN $h$
		with parameters $\bm{\theta}_0$, training dataset with domains $\mathcal{S}=\{ D_{i}\}_1^{\card{S}}$. Loss gradients $\bm{g}$. Learning rate $\eta$, perturbation parameter $\rho$, optimization step to start the annealing process $A_s$, optimization step to end the annealing process $A_e$, number of annealing iterations per optimization step $n_a$. Total number of training iterations $n$.
		\item[]
		\For{$t \leftarrow 1$ \textbf{to} $n$}
		\State  Sample a mini-batch: $\mathcal{B} \leftarrow \mathcal{B}_{\mathcal{D}_1}+...+\mathcal{B}_{\mathcal{D}_{\card{S}}}$
		\If{$A_s \leq t \leq A_e$}
		\State \texttt{\#\small{Begin Gradient-Guided Annealing:}}
		\State Compute the mini-batch loss at starting params: \\\hspace{1.5cm} $\mathcal{L}_B \leftarrow \mathcal{L}(\mathcal{B}; \bm{\theta}_t)$
		\State Calculate minimum domain grad pair sim: \\\hspace{1cm} 
		$\text{sim} \leftarrow \mathop{\min\limits_{\substack{
					i \leq \card{S},\, j \leq \card{S}\\
					i \neq j
		}}} 
		\left( \frac{\bm{g}^{T}_i \cdot \bm{g}_j}{\|\bm{g}_i\| \|\bm{g}_j\|} \right)$ 
		\State $\bm{\theta}_{t} \leftarrow \bm{\theta}_{t-1}, \bm{\theta}'_{t} \leftarrow \bm{\theta}_{t-1}$
		\For{ step $a \leftarrow 1$ \textbf{to} $n_a$}
		\State $\bm{\theta}_a \leftarrow \bm{\theta}_t + \mathcal{U}(-\rho, \rho)$
		\State Calculate minimum domain grad pair sim: \\\hspace{1.5cm} $\text{sim}_a \leftarrow
		\mathop{\min\limits_{\substack{
					i \leq \card{S},\, j \leq \card{S}\\
					i \neq j
		}}} 
		\left( \frac{\bm{g}^{T}_i \cdot \bm{g}_j}{\|\bm{g}_i\| \|\bm{g}_j\|} \right)$
		\State Compute the mini-batch loss at new params: \\\hspace{1.5cm} $\mathcal{L}_a \leftarrow \mathcal{L}(\mathcal{B}; \bm{\theta}_a)$
		\If{$\big(\text{sim}_a > \text{sim} \big) \wedge \big(\mathcal{L}_a - \mathcal{L}_{\mathcal{B}} < 0.1 \big) $}
		\State $\mathcal{L}_{\mathcal{B}} \leftarrow \mathcal{L}_a$, $\bm{\theta}'_t \leftarrow \bm{\theta}_a$, $\text{sim} \leftarrow \text{sim}_a$
		\EndIf
		\EndFor
		\EndIf
		\State \texttt{\#\small{Update weights:}}
		\State Compute mini-batch loss: $\mathcal{L}_\mathcal{B} =\mathcal{L}(\mathcal{B};\bm{\theta}'_t)$
		\State $\bm{\theta}_{t+1}=\bm{\theta}'_{t} -\eta \cdot \nabla\mathcal{L}_\mathcal{B}(\mathcal{B};\bm{\theta}'_t)$
		\State $t=t+1$
		\EndFor
	\end{algorithmic}
	\label{alg:GGA}
\end{algorithm}

The analysis presented in the previous section shows that we can use the
agreement of gradients $\nabla\mathcal{L}_i$ as an indicator of domain
invariance, since the less the model $h(\cdot|\bm{\theta})$ depends on the domain $d$, the more similar the expected value of the gradient,
$\mathbb{E}[\nabla\mathcal{L}_d]$, will be (see Eq. \eqref{eq:expgrad}). Inspired by Simulated Annealing \cite{kirkpatrick1983optimization}, we achieve 
this by adding random noise to the parameters of the model in search for a point with high domain gradient agreement, as measured by the increase of the minimum gradient similarity across any pair of domains,

\begin{equation}
	\text{grad\_sim} = 
	\mathop{\min\limits_{\substack{
				i \leq \card{S},\, j \leq \card{S}\\
				i \neq j
	}}} 
	\left( \frac{\bm{g}^{T}_i \cdot \bm{g}_j}{\|\bm{g}_i\| \|\bm{g}_j\|} \right)
	\label{eq:gradsim}
\end{equation}

In more detail, as it is common in DG literature, we start with a pre-trained
model $f(\cdot, \bm{\theta}_0)$ and perform a small number of warmup training
steps. This ensures that the model approaches a region of the parameter space
that achieves a low loss for the target problem. We then calculate the minimum 
pairwise gradient similarity among source domains and begin searching the
neighborhood of the parameter space for points where both: (a) the domain 
gradients agree and (b) the loss are reduced\footnote{In our experiments, we relax the loss constraint and accept the new parameter set when the gradient similarity has increased and training loss has been reduced within a specific range (i.e $\mathcal{L} - \mathcal{L}' < 0.1$), where $\mathcal{L}'$ is the loss value at the updated set of parameters.}. This is implemented through 
iterative random perturbations $\bm{\theta}' \leftarrow \bm{\theta} + \mathcal{U}(-\rho, \rho)$, where $\mathcal{U}$ is the multivariate uniform 
distribution. Each new point is selected only if it simultaneously achieves 
higher gradient agreement and lower loss (thus improving the Pareto front). 
After a fixed number of iterations, the point that has achieved the highest 
minimum gradient agreement and the lowest loss is selected. This process is repeated after every training iteration step. After a number of optimization steps, we allow the model to be trained without parameter perturbations, following a standard SGD-based procedure, as usual. The GGA algorithm is presented in \textbf{Algorithm} \ref{alg:GGA}.

Since the iterative search for parameter sets with improved gradient similarity 
introduces an additional computational overhead to training, we propose an 
alternative method which directly anneals the gradients during optimization and adds a considerably lower computation cost, inspired by Stochastic Gradient Langevin Dynamics (SGLD) \cite{welling2011bayesian}. We call this alternative method \textbf{GGA-L} and provide further details in the Supplementary Material.
\begin{table*}[h]
	\centering
	\small
	\caption{{\bf Comparison of \textbf{GGA} with the ERM baseline}. The top out-of-domain accuracies on five domain generalization benchmarks averaged over three trials, are presented. For better comparison with the results in Table 2, we report the average over PACS, VLCS, OfficeHome and TerraInc in the ``Avg.' column, while the average over all 5 datasets in the final one.}
	\label{table:baseline-results}
	\begin{tabular}{l|cccc|c||c|c}
		\toprule
		Algorithm & PACS & VLCS & OfficeHome & {TerraInc} &
		 Avg. & {DomainNet} & {Total Avg.}  \\
		\midrule
		ERM  & 
		85.5\scriptsize$\pm0.2$             & 
		77.3\scriptsize$\pm0.4$             & 
		\underline{66.5}\scriptsize$\pm0.3$             & 
		46.1\scriptsize$\pm1.8$             & 
		 68.9 & 
		43.8\scriptsize$\pm0.3$             & 63.9  \\
		\midrule
		\textbf{GGA}                & 
		\textbf{86.4}\scriptsize{$\pm0.5$}           & 
		\textbf{78.7}\scriptsize{$\pm0.8$}           & 
		\textbf{67.0}\scriptsize{$\pm0.3$}           &  
		\textbf{48.5}\scriptsize{$\pm1.1$}           & 
		\textbf{70.2}                                &
		\textbf{44.4}\scriptsize{$\pm0.2$}           & 
		\textbf{65.0}  \\
		

		\bottomrule
		
	\end{tabular}
\end{table*}

\begin{table*}[h]
	\centering
	\small
	\caption{\textbf{Comparison with state-of-the-art domain generalization methods.} Out-of-domain accuracies on five domain generalization benchmarks are shown. The results marked by $\dagger, \ddagger$ are copied from Gulrajani and Lopez-Paz \cite{gulrajani2020domainbed} and Wang et al. \cite{wang2023sharpness}, respectively. For fair comparison, the training of each algorithm combined with \textbf{GGA}, were run on the respective codebases. Average accuracies and standard errors are calculated from three trials for all experiments with GGA. In {\color{green} green} and {\color{red} red}, we highlight the performance boost and decrease of applying GGA on top of each algorithm respectively, averaged over three trials. In the ``Avg.'' column we report the average accuracy over the PACS, VLCS, OfficeHome and TerraInc. Due to computational resources, for DomainNet we do not combine GGA with previous methods and report the average over the 5 datasets in the final column.}
	\label{table:total-results}
	\renewcommand{\arraystretch}{1.1}
	\setlength{\tabcolsep}{3pt}
	\setlength{\abovetopsep}{0.5em}
	\begin{tabular}{l|cccc|c||c|c}
		\toprule
		Algorithm & PACS          & VLCS          & OfficeHome    & {TerraInc}    & {Avg.} & DomainNet & Total Avg. \\
		\midrule
		
		Mixstyle$^\ddagger$ \cite{zhou2021mixstyle}     & 
		85.2\scriptsize{$\pm0.3$} 
		\small{({\color{green}$+1.4$})}         & 
		77.9\scriptsize{$\pm0.5$} \small{({\color{green}$+0.6$})}          & 60.4\scriptsize{$\pm0.3$} \small{({\color{green}$+0.5$})}          & 44.0\scriptsize{$\pm0.7$}
		\small{({\color{green}$+4.0$})}          & 
		66.9 \small{({\color{green}$+1.6$})}         &
		34.0\scriptsize{$\pm0.1$}           & 60.3 \\    
		
		GroupDRO$^\ddagger$ \cite{Sagawa2020GroupDRO}    & 
		84.4\scriptsize{$\pm0.8$}          
		\small{({\color{green}$+1.6$})}   & 
		76.7\scriptsize{$\pm0.6$}          
		\small{({\color{green}$+0.6$})}   & 
		66.0\scriptsize{$\pm0.7$}             
		\small{({\color{red}$-0.4$})}   & 
		43.2\scriptsize{$\pm1.1$}              
		\small{({\color{green}$+5.2$})}   & 
		67.6 \small{({\color{green}$+1.7$})}         &
		33.3\scriptsize{$\pm0.2$}          & 60.7 \\      
		
		MMD$^\ddagger$ \cite{li2018mmd}                  & 
		84.7\scriptsize{$\pm0.5$}          
		\small{({\color{green}$+0.8$})} & 
		77.5\scriptsize{$\pm0.9$}            
		\small{({\color{green}$+0.6$})} & 
		66.3\scriptsize{$\pm0.1$}          
		\small{({\color{green}$+0.4$})} & 
		42.2\scriptsize{$\pm1.6$}            
		\small{({\color{green}$+5.1$})} & 
		67.7 \small{({\color{green}$+1.7$})} &      
		23.4\scriptsize{$\pm9.5$}           &  58.8 \\ 
		
		AND-mask \cite{shahtalebi2021sand} & 
		84.4\scriptsize{$\pm0.9$}  
		\small{({\color{green}$+0.1$})}& 
		78.1\scriptsize{$\pm0.9$}  
		\small{({\color{green}$+0.3$})}& 
		65.6\scriptsize{$\pm0.4$}  
		\small{({\color{green}$+1.2$})} & 
		44.6\scriptsize{$\pm0.3$} 
		\small{({\color{red}$-0.4$})}
		& 68.2 \small{({\color{green}$+0.3$})}    &
		37.2\scriptsize{$\pm0.6$} & 62.0 \\			
		
		ARM$^\ddagger$ \cite{zhang2020arm}        & 
		85.1\scriptsize{$\pm0.4$}          
		\small{({\color{green}$+0.1$})}& 
		77.6\scriptsize{$\pm0.3$} 
		\small{({\color{green}$+0.4$})}           & 
		64.8\scriptsize{$\pm0.3$} 
		\small{({\color{green}$+1.2$})}           & 
		45.5\scriptsize{$\pm0.3$} 
		\small{({\color{green}$+2.4$})}           & 
		68.3 \small{({\color{green}$+1.0$})}         &
		35.5\scriptsize{$\pm0.2$}          & 61.7 \\    
		
		IRM$^\dagger$ \cite{arjovsky2019irm}            & 
		83.5\scriptsize{$\pm0.8$}          
		\small{({\color{red}$-1.5$})}& 
		78.5\scriptsize{$\pm0.5$}          
		\small{({\color{red}$-0.9$})} & 
		64.3\scriptsize{$\pm2.2$}          
		\small{({\color{red}$-2.1$})}& 
		47.6\scriptsize{$\pm0.8$}          
		\small{({\color{red}$-3.9$})} & 
		68.5 \small{({\color{red}$-2.1$})}        &
		33.9\scriptsize{$\pm2.8$}          & 61.6 \\    
		
		MTL$^\ddagger$ \cite{blanchard2021mtl_marginal_transfer_learning}    & 
		84.6\scriptsize{$\pm0.5$}           
		\small{({\color{green}$+1.8$})} & 
		77.2\scriptsize{$\pm0.4$}          
		\small{({\color{green}$+0.9$})} & 
		66.4\scriptsize{$\pm0.5$}          
		\small{({\color{red}$-0.3$})} & 
		45.6\scriptsize{$\pm1.2$}          
		\small{({\color{green}$+0.8$})}&  
		68.5 \small{({\color{green}$+0.8$})}        &
		40.6\scriptsize{$\pm0.1$}          & 62.9 \\   

		VREx$^\ddagger$ \cite{krueger2020vrex}           & 
		84.9\scriptsize{$\pm0.6$} 
		\small{({\color{green}$+0.6$})}           & 
		78.3\scriptsize{$\pm0.2$} 
		\small{({\color{red}$-0.2$})}           & 
		66.4\scriptsize{$\pm0.6$} 
		\small{({\color{green}$+1.5$})}           & 
		46.4\scriptsize{$\pm0.6$} 
		\small{({\color{red}$-0.7$})}           & 
		69.0 \small{({\color{green}$+0.3$})}      &
		33.6\scriptsize{$\pm2.9$}          & 61.9   \\  
		
		MLDG$^\dagger$ \cite{li2018learning}                & 
		84.9\scriptsize{$\pm1.0$}          
		\small{({\color{green}$+1.2$})} &
		77.2\scriptsize{$\pm0.4$}          
		\small{({\color{green}$+1.3$})} & 
		66.8\scriptsize{$\pm0.6$}          
		\small{({\color{red}$-0.2$})} &
		47.7\scriptsize{$\pm0.2$}          
		\small{({\color{green}$+0.2$})} & 
		69.2 \small{({\color{green}$+0.6$})}    &
		41.2\scriptsize{$\pm0.1$}            & 63.6  \\  
		
		Mixup$^\dagger$ \cite{xu2020interdomain_mixup_aaai}             & 
		84.6\scriptsize{$\pm0.6$}            
		\small{({\color{green}$+1.0$})} & 
		77.4\scriptsize{$\pm0.6$}          
		\small{({\color{green}$+0.4$})} & 
		68.1\scriptsize{$\pm0.3$}            
		\small{({\color{green}$+0.1$})} & 
		47.9\scriptsize{$\pm0.8$}            
		\small{({\color{green}$+0.5$})} & 
		69.5   \small{({\color{green}$+0.5$})}    &
		39.2\scriptsize{$\pm0.1$}            & 63.4    \\ 

		SagNet$^\dagger$ \cite{nam2019sagnet}           &             
		86.3\scriptsize{$\pm0.2$} 
		\small{({\color{red}$-1.0$})} & 
		77.8\scriptsize{$\pm0.5$}          
		\small{({\color{green}$+0.9$})} & 
		68.1\scriptsize{$\pm0.1$}          
		\small{({\color{green}$+0.3$})}& 
		48.6\scriptsize{$\pm1.0$}          
		\small{({\color{green}$+0.7$})}& 
		70.2 \small{({\color{green}$+0.4$})}        &
		40.3\scriptsize{$\pm0.1$}          &  64.2 \\    

		CORAL$^\dagger$ \cite{sun2016coral}             & 
		86.2\scriptsize{$\pm0.3$}          
		\small{({\color{green}$+0.7$})}& 78.8\scriptsize{$\pm0.6$}          
		\small{({\color{red}$-0.4$})} & 
		68.7\scriptsize{$\pm0.3$}          
		\small{({\color{green}$+0.2$})} & 
		47.6\scriptsize{$\pm1.0$}          
		\small{({\color{green}$+0.3$})}& 
		70.3    \small{({\color{green}$+0.2$})}    &
		41.5\scriptsize{$\pm0.1$}          & 64.5   \\   
		
		\midrule
		
		RSC$^\dagger$ \cite{huang2020rsc}               & 
		85.2\scriptsize{$\pm0.9$}          
		\small{({\color{green}$+0.1$})}& 77.1\scriptsize{$\pm0.5$}          
		\small{({\color{green}$+0.2$})}& 
		65.5\scriptsize{$\pm0.9$} \small{($+0.0$)}        & 
		46.6\scriptsize{$\pm1.0$}          
		\small{({\color{green}$+0.2$})}& 
		68.6 \small{({\color{green}$+0.1$})}   &
		38.9\scriptsize{$\pm0.5$}          & 62.7 \\    
		
		Fish $^\ddagger$ \cite{shi2021gradient}                     & 
		85.5\scriptsize{$\pm0.3$}          
		\small{({\color{green}$+0.1$})} & 
		77.8\scriptsize{$\pm0.3$}          
		\small{({\color{green}$+0.9$})} & 
		68.6\scriptsize{$\pm0.4$}            
		\small{({\color{red}$-0.6$})} & 
		45.1\scriptsize{$\pm1.3$}            
		\small{({\color{green}$+1.8$})} & 
		69.3 \small{({\color{green}$+0.6$})}      &
		42.7\scriptsize{$\pm0.2$}            &   63.9    \\  
		
		SAM $^\ddagger$ \cite{foret2020sharpness}  & 
		85.8\scriptsize$\pm0.2$             
		\small{({\color{green}$+0.3$})} & 
		79.4\scriptsize$\pm0.1$              
		\small{({\color{green}$+0.2$})} & 
		{69.6}\scriptsize$\pm0.1$              
		\small{({\color{red}$-1.0$})} & 
		43.3\scriptsize$\pm0.7$              
		\small{({\color{green}$+2.9$})} & 
		69.5 \small{({\color{green}$+0.6$})}   &
		44.3\scriptsize$\pm0.0$              &  64.5 \\   
		
		GSAM $^\ddagger$ \cite{zhuang2022surrogate}  & 
		85.9\scriptsize$\pm0.1$           
		\small{({\color{green}$+0.6$})} & 
		79.1\scriptsize$\pm0.2$             
		\small{({\color{black}$\pm0.0$})} & 
		69.3\scriptsize$\pm0.0$             
		\small{({\color{red}$-0.7$})}& 
		47.0\scriptsize$\pm0.8$             
		\small{({\color{red}$-1.0$})}& 
		70.3  \small{({\color{red}$-0.3$})}  &
		44.6\scriptsize$\pm0.2$             &   65.1 \\
		
		SAGM $^\ddagger$ \cite{wang2023sharpness}       & 
		{86.6}\scriptsize{$\pm0.2$}           
		\small{({\color{green}$+0.2$})}& 
		{80.0}\scriptsize{$\pm0.3$}           
		\small{({\color{black}$\pm0.0$})}& 
		{70.1}\scriptsize{$\pm0.2$}           
		\small{({\color{red}$-0.6$})}& 
		{48.8}\scriptsize{$\pm0.9$}           
		\small{({\color{red}$-0.1$})}& 
		{71.4} \small{({\color{red}$-0.1$})}  &
		45.0\scriptsize{$\pm0.2$}           &  66.1 \\
		

		\bottomrule
		
	\end{tabular}
\end{table*}

\section{Experiments}
\label{sec:experiments}

\subsection{Experimental setup and implementation details}

In our experiments, we follow the protocol of DomainBed 
\cite{gulrajani2020domainbed} and exhaustively evaluate our algorithm against state-of-the-art algorithms on five DG benchmarks, using the same dataset splits and model selection. For the hyperparameter search space, we follow \cite{cha2021swad} in order to avoid the high computational burden of DomainBed. The datasets included in the benchmarks are, PACS \cite{li2017deeper} (9,991 images, 7 classes, and 4 domains), VLCS \cite{fang2013unbiased} (10,729 images, 5 classes, and 4 domains), OfficeHome \cite{venkateswara2017deep} (15,588 images, 65 classes, and 4 domains), TerraIncognita \cite{beery2018recognition} (24,788 images, 10 classes, and 4 domains) and DomainNet \cite{peng2019moment} (586,575 images, 345 classes,
and 6 domains).

In all experiments, the \textit{leave-one-domain-out} cross-validation protocol is followed. Specifically, in each run a single domain is left out 
as the target (test) domain, while the rest of the domains are used for 
training. The final performance of each algorithm is calculated by averaging the top-1 accuracy on the target domain, with different train-validation splits. For training, we utilize a ResNet-50 
\cite{he2016deep} pretrained on ImageNet \cite{russakovsky2015imagenet} for the backbone feature extractor and ADAM for the optimizer. Regarding GGA, 
during training we let each algorithm run for several training iterations depending on the dataset and then begin the parameter space 
search, as described in Section \ref{sec:methods}. For the neighborhood size
in the search process, we set $\rho$ to $1e-6$ for VLCS and $1e-5$ for all other datasets\footnote{The selection of $\rho$ was based experimentation between values close to or larger than the learning rate during training. The sensitivity analysis for the PACS dataset presented in subsection \ref*{sensitivity} also reveals that $\rho = 1e-5$ yields the best performance.} and search for a total of $A = 250$ steps before moving to the next mini-batch from each domain, in each dataset. In all experiments, we perform search iterations for $100$ different mini-batches during early training stages. The rest of the hyperparameters, such as learning rate, weight decay and dropout rate, are tuned according to \cite{cha2021swad} and are presented in Table \ref{table:hyperparameters}.
To account for the variability introduced in the random search of GGA, we 
repeat the experiments with 3 different seeds for each dataset. All models were trained on a cluster containing $4\times40$GB NVIDIA A$100$ GPU cards, split into $8$ $20$GB virtual MIG devices and $1\times24$GB NVIDIA RTX A$5000$ GPU card.

\begin{table}[h]
	\centering
		\resizebox{\linewidth}{!}{
		\begin{tabular}{lrrrrr} 
			\toprule
			\textbf{Hyperparameter}& PACS & VLCS  & OH & TI & DN\\ \midrule
			Learning rate&{3e-5}&{1e-5}&{3e-5}& 3e-5 & 3e-5 \\ 
			Dropout & 0.5 & 0.5 & 0.5 & 0.5 & 0.5 \\ 
			Weight decay&1e-4 & 1e-4&1e-4 & 1e-4 & 1e-4 \\ 
			Training Steps & 5000 & 5000 & 5000 & 5000 & 15000 \\
			$\rho$ & {1e-5} & {1e-6} & {1e-5} & {1e-5} & {1e-5} \\
			GGA Start-End & 100-200 & 100-200 & 100-200 & 100-200 & 100-200 \\ 
			\bottomrule
		\end{tabular}
				}
	\caption{Hyperparameters for DG experiments. OH, TI and DN stand for OfficeHome, TerraIncognita and DomainNet respectively. $\rho$ is the parameter space search used during the annealing steps in GGA, while the last row indicates the training iterations during which GGA occurs.}
	\label{table:hyperparameters}
\end{table}

\subsection{Comparative Evaluation}
\label{results}

The average OOD performances of the baseline vanilla ERM \cite{vapnik1998statistical} and state-of-the-art DG methods on a total of 5 DG benchmarks, are reported in Tables \ref{table:baseline-results} and \ref{table:total-results} respectively. The results for each separate domain in each benchmark are reported in the Appendix. In the experiments, we apply GGA on-top of the rest of the DG algorithms and show that its properties generalize to other methods as well, boosting their overall performance in most cases. It should also be noted that GGA can be adapted to training scenarios where distributions shifts are present in training data. 

Initially, to validate whether parameter annealing in the early stages of 
model training leads to models with increased generalizabilty, we compare the 
results with the vanilla ERM baseline. As presented in Table \ref{table:baseline-results}, GGA is able to boost the performance of the baseline model by an average of 1.1\% on all benchmarks, demonstrating their efficacy. What's more, as evident from the results in Table \ref{table:total-results}, {GGA} is on average competitive with the SoTA across all 5 datasets. For further evaluation, we also compare GGA with state-of-the-art DG methods \cite{li2018mmd, zhou2021mixstyle, Sagawa2020GroupDRO, arjovsky2019irm, zhang2020arm, krueger2020vrex, shahtalebi2021sand, ganin2016dann, huang2020rsc, blanchard2021mtl_marginal_transfer_learning, xu2020interdomain_mixup_aaai, li2018learning, shi2021gradient, nam2019sagnet, sun2016coral, foret2020sharpness, zhuang2022surrogate, wang2023sharpness, vapnik1998statistical}, before applying GGA to them as well. We note that we only include previous works that have implemented a ResNet-50 as the backbone encoder and do not use additional components or ensembles. In Table \ref{table:total-results} we differentiate between the methods that operate 
on model gradients \cite{huang2020rsc, shi2021gradient, foret2020sharpness, zhuang2022surrogate, wang2023sharpness} and the rest of the algorithms. Even without its application to other methods, GGA is able to surpass most of the previously proposed algorithms and remain competitive with the state-of-the-art. What's more important is that the application of GGA in conjunction with the rest of the DG methods, proves beneficial and ultimately boosts their overall performance in almost each case by an average of around $1\%$ and in some cases up to even $5.2\%$. With regards to IRM, which seems to be significantly impacted negatively by the application of GGA, its learning objective emphasizes on simultaneously minimizing the training loss of each source domain and not necessarily on the pairwise agreement of gradients among domains. It is therefore not able to converge to a good solution after the initial model's weights have been perturbed during training.

From the experimental results, it is evident that the application of GGA 
and its search for parameter values where gradients align between domains is beneficial to model training. By introducing the proposed annealing step before the final stages of  training, the majority of the models exceed their previous performance and exhibit improved generalization capabilities.

\begin{figure*}[t]
	\centering
	\includegraphics[width=\textwidth]{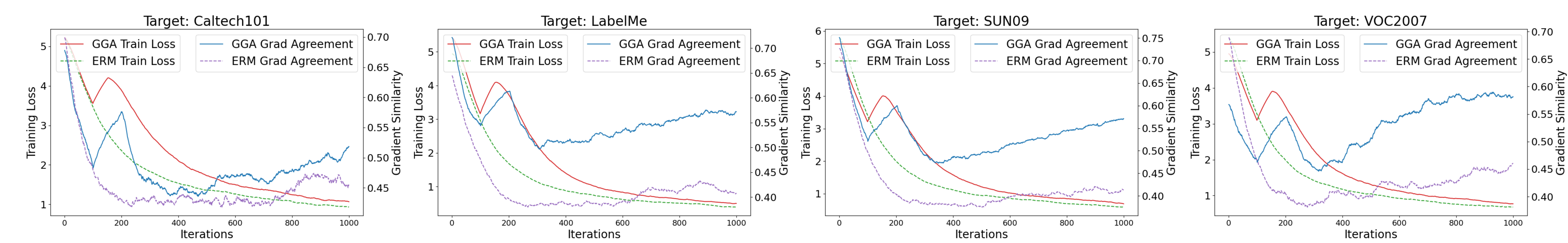}
	\caption{Impact of GGA on gradient alignment during model training on the VLCS dataset. As illustrated, in the case of vanilla ERM, even though the training loss is minimized, the average cosine similarity among domain gradients remains low. In the case of GGA however, after the algorithm searches for points in the parameter space with increased gradient similarity, the gradients continue to agree during training, while the total training loss is also minimized.}
	\label{vlcs-grads}
\end{figure*}

\subsection{Evaluating the impact of GGA on gradient disagreement}
\label{gga-evaluation}

As discussed in Section \ref{sec:methods}, when a training dataset is composed
as a mixture of multiple domains, conflicting gradients between mini-batches drawn
from each domain lead to models that do not infer based on domain-invariant
features and which generalize to previously unseen data samples, but are hindered
by domain-specific, spurious correlations. This is evident in the case of
vanilla models trained via ERM where the average gradient similarity among
domains continues to remain low upon reaching a local minima. Our hypothesis is
that this behavior can be avoided by searching for a parameter set of common
agreement between domains before optimizing via gradient descent.

To demonstrate the operation of the proposed algorithm in practice against ERM, we calculate the average gradient cosine similarity between mini-batches from source domains during training for the VLCS dataset, along 
with the training loss in each iteration. As a result, each sub-figure in Figure
\ref{vlcs-grads} illustrates the progression of the training gradient alignment
between domains, against the total training loss. 

As expected, in the very initial iterations the gradients of the pretrained
model parameters point towards a common direction. However, in the case of ERM
as training progresses and the loss is minimized, the domain gradients begin to
disagree leading the model to converge to undesirable minima that do not
generalize across domains. On the other hand, when GGA is applied the model
searches for parameters such that gradients are aligned before continuing
training. This is illustrated by the spike in gradient similarity,
during iterations $100$ up to $200$. After GGA concludes, we observe that the
model continues training by descending into minima where gradients agree among
domains.

\subsection{Sensitivity Analysis}
\label{sensitivity}

\begin{figure}[t]
	\centering
	\includegraphics[width=\columnwidth]{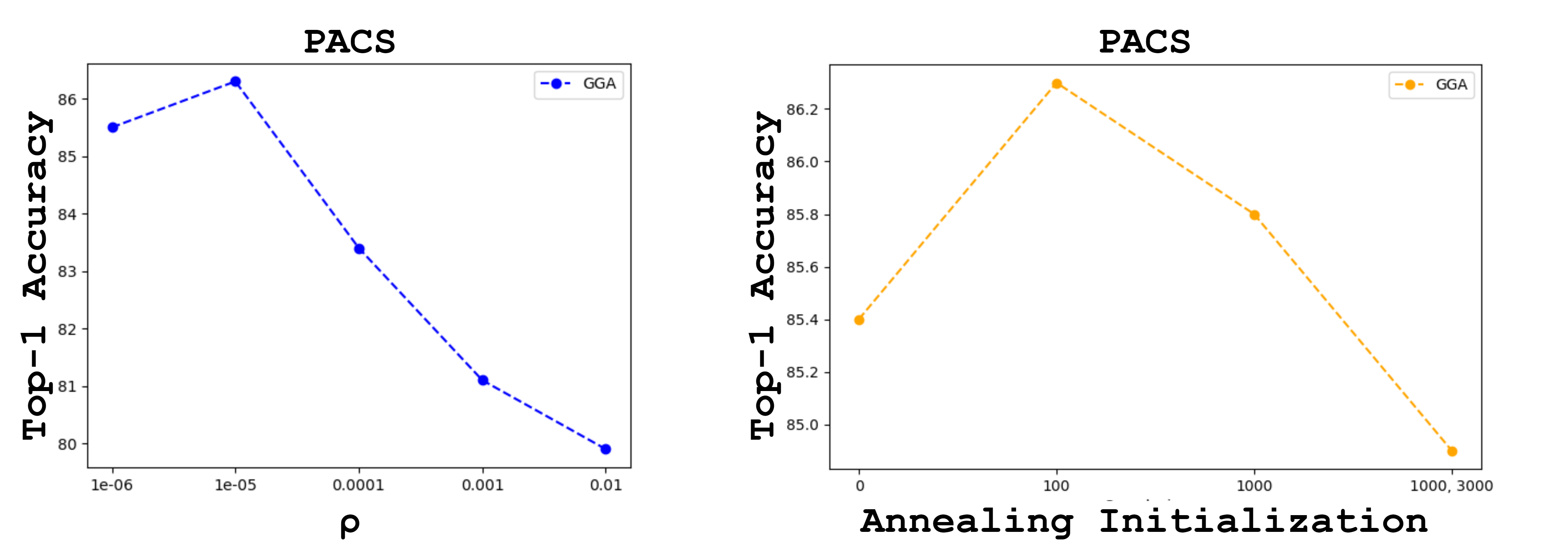}
\caption{Sensitivity analysis of the parameter space search magnitude $\rho$ and the training stage application of GGA. The analysis on PACS reveals smaller weight perturbations and gradient annealing during early training iterations lead to increased model performance.}
\label{sensitivity-fig}
\end{figure}

The two core parameters of GGA, are the size of the parameter search $\rho$ and
the moment of our methods implementation during training, i.e., during the
early, mid or late training stages. To justify their selection, we conduct a
sensitivity analysis (Fig. \ref{sensitivity-fig}) by varying one of the above
parameters while fixing the other at its optimal value.

Regarding the magnitude of weight perturbations during the application of GGA,
we found that the optimal value was $\rho = 1e-5$. As illustrated in Figure
\ref{sensitivity-fig}, a larger magnitude of perturbation led to decreased model
performance. Intuitively, the application of larger noise to the model
parameters leads to sets that are not close to the solution, making it
increasingly difficult for the model to converge. On the other hand, smaller
perturbations seem to have little to no effect on training, as the search is
limited to spaces near the current parameters, which is why the model
performance falls back close to that of ERM. With regards to the stage of
training during which GGA will be applied, we found that the models yielded
better performance when the search was initialized in earlier stages.We hypothesize that applying perturbations near the end of training displaces the model from a local optimum, requiring additional iterations to converge.

\section{Conclusions}
\label{sec:conclusions}

In this work, we investigated gradient conflicts in the context of domain generalization, leading to the development of GGA; a simple yet effective algorithm that identifies points in the parameter space where domain gradients align early in training before continuing optimization, ultimately enhancing model generalization. Through a comprehensive comparative analysis we have demonstrated the efficacy of GGA, which is able to achieve superior generalization capabilities in standard DG benchmarks and outperform SoTA 
baselines. What's more, as GGA is both model and method agnostic it can also be utilized by other algorithms. Interestingly, its combination with previous methods proves beneficial, as their majority yields improved performance over all benchmarks. Finally, we validate the impact of our algorithm on gradient alignment through additional experiments, showing that, unlike the baseline, domain gradients align during training.

However, our method does not come without some key limitations. First of all,
the annealing steps and gradient similarity computation add computational overhead. Furthermore, GGA relies on domain labels, making it inapplicable to single-source DG, where domain labels are not available during training. Its effectiveness is also batch-size dependent, as larger batches provide further information regarding gradient alignment and can lead to improved results. Future work will further explore gradient alignment in multi-domain settings and refine GGA to make annealing steps adaptive rather than random.
\\

\noindent\textbf{Acknowledgement.} This work was supported by the European Union’s Horizon 2020 research and innovation programme under Grant Agreement No. 965231, project REBECCA (REsearch on BrEast Cancer induced chronic conditions supported by Causal Analysis of multi-source
data).

{
    \small
    \bibliographystyle{ieeenat_fullname}
    \bibliography{main}
}


\clearpage
\setcounter{page}{1}
\maketitlesupplementary

The following materials are provided in this supplementary file:
\begin{itemize}
	\item An extended literature review discussion, helpful for navigating the Domain Generalization literature under the scope of computer vision.
	\item A computational analysis regarding the application of GGA.
	\item Detailed results for each dataset domain and algorithm, presented in Table 2 of the main text, along with extra experiments on the ColoredMNIST and RotatedMNIST datasets.
\end{itemize}

\section{Extended Literature Review}
\label{sec:extended-lit-review}

Domain Generalization (DG) \cite{wang2022generalizing} is arguably one of the 
most difficult and fundamental problems of Machine Learning (ML) today. 
Unsurprisingly, a vast number of researchers have poured effort into advancing the field, where findings have been applied to various areas, such 
as Natural Language Processing \cite{hupkes2023taxonomy}, Reinforcement Learning \cite{li2018learning}, Healthcare and Medicine \cite{9298838, 10233054}, Time-Series forecasting \cite{du2021adarnn}, Fault Diagnosis 
\cite{9174912} and, of course, Computer Vision \cite{wang2022generalizing}. 
Even though not covering the entire field of DG, this section aims to present 
a taxonomy of the general DG methodologies developed in CV, for producing 
robust models that can generalize to previously unseen data, and attempts to 
assist potential readers navigate the past literature, while also 
categorizing our proposed method among its predecessors. 
Domain Generalization methods can be categorized into three major groups, depending on their operation during the process of model training, namely: 
(a) Data Manipulation, (b) Representation Learning and, (c) Learning 
Algorithm. Furthermore, as mentioned in the main text, DG algorithms can 
either leverage domain labels during training (multi-source), or completely
disregard the knowledge of existing domain shifts in their training data and 
handle them as a single distribution (single-source). 

\textbf{Data Manipulation}. As its name suggests, methods 
included in this group focus on either perturbing existing samples (\textit{data augmentation}) or creating novel ones (\textit{data generation}), 
in order to regularize the training of machine learning models, avoid 
overfitting and improve their generalizability. The basic idea in data 
manipulation methodologies is to simulate domain shift by creating diverse 
data samples, which can in turn mimic the entirety of distributions present 
in the input space. Regarding data augmentation, most popular techniques 
include traditional image transformations, such as random flip, rotation and 
color distortion. Even though these augmentations can be randomly applied 
during training, without needing domain labels, it has been shown that their 
selection significantly affects model performance. For example, the 
authors of \cite{volpi2019addressing} define novel augmentation rules that 
push the perturbed images to diverge as much as possible from the original 
ones. Additionally, image augmentations prove effective towards overcoming 
domain shifts in medical image classification \cite{otalora2019staining, zhang2020generalizing}, where transformations can replicate shifts caused by 
the use of different devices. On the other hand, multiple data augmentation 
methods were also inspired by adversarial attacks and use adversarial gradients to distort the input images \cite{volpi2018generalizing, qiao2020learning}, or use randomly initialized convolutional networks for transforming samples \cite{choi2023progressive}. These techniques act as regularizers during model training, allowing them to learn generalizable image representations. The generation of novel data domains is also a well 
researched area in the data manipulation group. In addition to using domain gradients for synthesizing novel domains \cite{shankar2018generalizing}, 
several methods took advantage of style transfer \cite{huang2017arbitrary} 
and either map the styles of images to that existing source domains \cite{borlino2021rethinking} or create novel styles \cite{yue2019domain}. On a similar note, mixing the styles of training images by conventional methods  \cite{xu2020interdomain_mixup_aaai, zhou2021mixstyle} or with the generative models \cite{wang2024multi} also proves beneficial.

\textbf{Representation Learning}. This group of methods is arguably the most 
prominent in DG and has been the central focus of ML \cite{6472238}. 
Following the formulation in the main text, given a labeling function $h$ 
that maps input observations $\bm{x}$ to their labels $y$, we can decompose it into $h = f \circ g$, where $g$ is a parametric function that learns 
representations of $x$ and $f$ is the classifier function. The goal of 
representation learning can be summarized as follows:

\begin{equation}
	\min_{f, g} \mathbb{E}_{x,y} \ell(f(g(\bm{x}; \bm{\theta})), y) +\lambda\ell_\text{reg}
\end{equation}
where $\ell$ the loss function to be minimized and $\ell_\text{reg}$ a 
regularizer. Methods included in this group, focus on learning a robust and generalizable representation learning function $g$. The algorithms included 
in this group can be further categorized into three sub-groups. \textit{Feature disentanglement} \cite{Zhang_2022_CVPR} methods intend to extract disentangled feature vectors from samples, where each dimension can be linked to a subset of data generating factors. The main idea is to produce a model that extracts a representation that can be further decomposed into domain-specific, domain-invariant, and class-specific features. To that end, the authors of \cite{piratla2020efficient} present CSD, which jointly learns a domain-invariant and domain-specific component in the final embedding and enables the extraction of disentangled representations, whereas the authors of \cite{chattopadhyay2020learning} propose learning domain specific 
masks during training to improve model robustness. Generative models have also been proposed in the disentangled representation learning literature for DG, with variational autoencoders (VAEs) and GANs \cite{chen2016infogan} being utilized for learning distinct latent subspaces for class- and domain-specific features \cite{ilse2020diva}. Another promising category of 
methods aiming to produce disentangled representations is that of 
Causality-Inspired algorithms. In causal representation learning, a domain 
shift can be thought of as an intervention, subsequentially leading the development of models that aim to uncover the true causal data generating factors. Naturally, the prediction of a model should not be affected by interventions on spuriously correlated but irrelevant features, such as the background, color or style of the image. Under this causal consideration, the authors of \cite{mahajan2021domain} propose a structural causal model in 
order to model within-class variations and leverage the fact that inputs 
across domains should have the same representation, given that they derive 
from the same object. Similar to disentangled representations, there have 
been proposed methods in the literature that focus on completely disregarding 
domain-relevant from the final feature vectors, deriving solely 
domain-invariant representations. Based on the initial findings of \cite{ben2006analysis}, numerous works have presented algorithms that aim to minimize the representation differences across multiple source domains within a specific feature space, ensuring they become domain invariant, ultimately enabling the trained model to effectively generalize to previously unseen domains. In one of the most notable previous 
works in this category, Arjovsky et al. \cite{arjovsky2019irm} enforce the 
optimal classifier on top of the representation space to be the same across domains and simultaneously minimize the loss across distributions. The above 
idea of Invariant Risk Minimization (IRM) has been extended to several other
works. For example, the authors of \cite{krueger2021out} propose minimizing 
the variance of source-domain risks, by minimizing their extrapolated risk, 
while the authors of \cite{zhang2020arm} propose adapting to domain shift and 
producing invariant representations. Finally, an alternative route towards
learning generalized representations is via regularization strategies. 
The most representative group of methods in this category is \textit{Gradient-Based operations}, which utilize gradient information during 
model training. In \cite{huang2020rsc}, the authors propose learning robust
representations by discarding the most dominant gradients in each training 
iteration under the assumption that they are correlated with domain-specific
features present in the source data. Another popular strategy is to seek for 
flat minima \cite{foret2020sharpness, cha2021swad} in the loss landscape of 
neural networks during training, assuming that models that converge to flat minima exhibit increased generalization capabilities \cite{zhuang2022surrogate, wang2023sharpness}. 
What's more, Shi et al. \cite{shi2021gradient} hypothesized that gradients 
among domains should match and proposed an approximation of a loss inducing the maximization of the gradient inner product during training. Our method (GGA) can be categorized in this group of gradient operations, as it considers
the similarity of domain gradients in the early iterations of model training
and seeks for sets in the parameter space with increased gradient alignment, before continuing the optimization procedure. 

\textbf{Learning Algorithm}. In addition to manipulating the input space or 
feature extractor, DG methods were also researched under the scope of 
alternative ML learning paradigms, such as \textit{ensemble}, \textit{meta}, \textit{domain-adversarial}, \textit{self-supervised} and 
\textit{reinforcement} learning. In this section we present the most exemplary
works in each category. \textit{Ensemble-Learning} in DG initially combined
several copies of the same network, each of which is trained on a specific 
domain \cite{zhou2021domain, ding2017deep}. Alternatively, instead of using several networks, \cite{yosinski2014transferable} proposed sharing shallow 
layers among CNNs. During inference, the final prediction is produced by 
either simple \cite{zhou2021domain} or weighted averaging 
\cite{wang2020dofe}. In \textit{Meta-Learning} for DG, Li et al. \cite{li2018learning} propose MLDG and split the source domains into 
meta-train and meta-test splits to mimic the effects of domain shift during 
training. Similarly, \cite{balaji2018metareg} proposes learning a meta regularizer for the classifier, while MAML \cite{finn2017model} was proposed
for improved parameter initialization. Another approach is that of \textit{Adversarial Learning} (AL). In the context of DG, the aim of 
adversarial learning is to train a classifier to distinguish between source domains \cite{matsuura2020domain} and ultimately learn domain-agnostic 
features from the samples that can be generalized to unseen data 
\cite{li2018deep}. Other learning paradigms such as \textit{Self-Supervised} 
learning have also been explored in DG, which leverages unlabeled data 
samples to derive generalized representations. Notably, the authors of 
\cite{carlucci2019domain} introduce a self-supervised jigsaw-solving puzzle 
task to push the model to learn robust representations. Furthermore, 
contrastive learning has also been shown to improve model performance. 
Specifically, SelfReg \cite{kim2021selfreg} utilizes self-supervised contrastive losses to bring latent representations of same-class samples closer. Similarly, the authors of \cite{ballas2024multi} introduce 
a contrastive loss for representations extracted from intermediate 
layers of the network. Finally, \textit{Reinforcement learning} has also been 
applied in the context of DG. Indicatively, previous works have explored randomizing the environments of an RL agent for transferring them to 
real-world scenarios \cite{tobin2017domain,lee2019network}, whereas \cite{laskin2020curl} researches the combination of RL with contrastive 
learning.

\section{Computational Analyis}
\label{computation}

\subsection{Experiment Infrastructure}
\label{experiment-infra}

Each and every experiment is conducted on a cluster containing $4\times40$GB NVIDIA A$100$ GPU cards, split into $8$ $20$GB virtual MIG devices and $1\times24$GB NVIDIA RTX A$5000$ GPU card, via a SLURM workload manager.

\subsection{Complexity Analysis}
\label{sec:complexity}

Each GGA training iteration includes computing model gradients $S\cdot n_a$ times for each training step, where $S$ is the number of source domains and $n_a$ is the number of search steps. These GGA training iterations only take place in the early stages of training and for a small percentage of the total training iterations (2\% in our experiments). The rest of the iterations are vanilla ERM. Furthermore, inference is not affected by the application of GGA 
during training.

\section{An Alternative Method for Gradient-Guided Annealing: GGA-L}
In this section we propose we propose an alternative method to GGA, 
which integrates noise directly into the gradient update step rather than modifying weights separately. The key idea is to inject controlled noise during vanilla optimization, guiding updates toward regions with more stable gradient similarity across domains. Normally, in SGD, parameters update as: 
$\bm{\theta}_{t+1}=\bm{\theta}_{t} -\eta \cdot \nabla\mathcal{L}_\mathcal{B}(\mathcal{B};\bm{\theta}_t)$ for a sample batch $\mathcal{B}$. Instead of perturbing model weights manually for specific training steps and checking gradient similarity iteratively, as in GGA, we propose injecting dynamic noise based on domain gradient similarity directly into the update step, as follows:

\begin{equation}
	\bm{\theta_{t+1}}=\bm{\theta_{t}} -\eta \cdot (\nabla\mathcal{L}_\mathcal{B}(\mathcal{B};\bm{\theta_t}) +
	\alpha \cdot \bm{\xi})
\end{equation}
\\

where $\bm{\xi} \sim \mathcal{U}(0,1)$ is noise drawn from the Uniform distribution and
\begin{equation}
	\alpha = \gamma \cdot (1 - \frac{2}{\card{S}(\card{S} - 1)} \sum_{\substack{i=1, j=1 \\ i \ne j}}^{\card{S}} \left( \frac{\bm{g}^{T}_i \cdot \bm{g}_j}{\|\bm{g}_i\| \|\bm{g}_j\|} \right))
\end{equation} 
is a dynamic scaling factor depending on the average gradient similarity between domains and $\gamma$ is a hyperparameter controlling the noise intensity. The noise term $\alpha$ acts as an implicit exploration, pushing the optimizer toward parameter sets where the gradient similarity is naturally higher. When gradients from different domains are already similar
(i.e., high cosine similarity), $\alpha$ remains small, whereas if the
gradients diverge, $\alpha$ increases, adding more noise to encourage exploration in the parameter space. As the above proposed method is inspired by Stochastic Gradient Langevin Dynamics (SGLD) \cite{welling2011bayesian}, we call it GGA-L.

Following the same experimental protocol and hyperparameters as in the main paper, we show the results on PACS, VLCS, OfficeHome and DomainNet in Table \ref{table:extra-gga-l-results}. We set the $\gamma$ hyperparameter to $1e-3$ 
for all experiments expect DomainNet where we set it to $1e-4$ and apply the noisy update rule throughout training. For batch sizes we use 32 for PACS, VLCS and DomainNet and 48 for OfficeHome and TerraIncognita. As apparent from the results, GGA-L is able to perform similarly to GGA but adds a considerably lower computational cost as the gradients for each domain are only calculated once per batch. GGA-L is presented in Algorithm \ref{alg:GGA-L}.

\begin{algorithm}[t!]
	\caption{Implementing GGA-L in training}
	\begin{algorithmic}[1]
	\Require   Pretrained DNN $h$
	with parameters $\bm{\theta}_0$, training dataset with domains $\mathcal{S}=\{ D_{i}\}_1^{\card{S}}$. Loss gradients $\bm{g}$. Learning rate $\eta$, noise $\xi$ drawn from the Uniform distribution  $\mathcal{U}(0,1)$, noise scaling factor $\alpha$, hyperparameter $\gamma$ controlling the noise intensity. Total number of training iterations $n$.
		\item[]
		\For{$t \leftarrow 1$ \textbf{to} $n$}
		\State  Sample a mini-batch: $\mathcal{B} \leftarrow \mathcal{B}_{\mathcal{D}_1}+...+\mathcal{B}_{\mathcal{D}_{\card{S}}}$
		\State Compute the mini-batch loss at starting params: \\\hspace{1.5cm} $\mathcal{L}_B \leftarrow \mathcal{L}(\mathcal{B}; \bm{\theta}_t)$
		\State \texttt{\#\small{GGA-L}}/
		\State Calculate minimum domain grad pair sim: \\\hspace{1cm} 
		$\text{sim} \leftarrow \displaystyle\frac{2}{\card{S}(\card{S} - 1)} \sum\limits_{\substack{i=1,\, j=1 \\ i \ne j}}^{\card{S}} \left( \frac{\bm{g}^{T}_i \cdot \bm{g}_j}{\|\bm{g}_i\| \|\bm{g}_j\|} \right)$
		\State Calculate noise scaling factor $\alpha$\\\hspace{1.5cm}
		$\alpha = \gamma \cdot (1 - \text{sim})$
		\State \texttt{\#\small{Update weights:}}
		\State $\bm{\theta}_{t+1}=\bm{\theta}_{t} -\eta \cdot \nabla(\mathcal{L}_\mathcal{B}(\mathcal{B};\bm{\theta}_t) + \alpha \cdot \bm{\xi})$
		\State $t=t+1$
		\EndFor
	\end{algorithmic}
	\label{alg:GGA-L}
\end{algorithm}

\begin{table*}[h]
	\centering
	\small
	\caption{{\bf Comparison of \textbf{GGA} and \textbf{GGA-L} with the ERM baseline}.The top out-of-domain accuracies on five domain generalization benchmarks averaged over three trials, are presented. We report the average over PACS, VLCS, OfficeHome and TerraInc in the ``Avg.' column, while the average over all 5 datasets in the final one.}
	\label{table:extra-gga-l-results}
	\begin{tabular}{l|cccc|c||c|c}
		\toprule
		Algorithm & PACS & VLCS & OfficeHome & {TerraInc} & {Avg.}& {DomainNet} & {Total Avg.}  \\
		\midrule
		ERM  & 
		85.5\scriptsize$\pm0.2$             & 
		77.3\scriptsize$\pm0.4$             & 
		66.5\scriptsize$\pm0.3$             & 
		46.1\scriptsize$\pm1.8$             & 
		68.9 &
		43.8\scriptsize$\pm0.3$             & 63.9  \\
		\midrule
		{GGA}                & 
		\underline{86.4}\scriptsize{$\pm0.5$}           & 
		\textbf{78.7}\scriptsize{$\pm0.8$}           & 
		\textbf{67.0}\scriptsize{$\pm0.3$}           &  
		\underline{48.5}\scriptsize{$\pm1.1$}           &
		\underline{70.2} & 
		\textbf{44.5}\scriptsize{$\pm0.2$}           & 
		\underline{65.0}  \\
		
		\textbf{GGA-L}                & 
		\textbf{86.5}\scriptsize{$\pm0.7$}           & 
		\underline{78.4}\scriptsize{$\pm0.9$}           & 
		\underline{66.5}\scriptsize{$\pm0.4$}           &  
		\textbf{49.8}\scriptsize{$\pm1.8$}           &
		\textbf{70.3} &
		\textbf{44.5}\scriptsize{$\pm0.2$}           & 
		\textbf{65.1}  \\

		\bottomrule
		
	\end{tabular}
\end{table*}

\section{Full Experimental Results}
\label{sec:full-results}
In this section, we show detailed results of Table 2 in the main text.
The results marked by $\dagger, \ddagger$ are copied from Gulrajani and
Lopez-Paz \cite{gulrajani2020domainbed} and Wang \etal 
\cite{wang2023sharpness}, respectively. Standard errors for the baseline methods are reported from three trials, if available from past literature.
In {\color{green} green} and {\color{red} red}, we highlight the performance boost and decrease of applying \textbf{GGA} on top of each algorithm respectively, averaged over three trials. In addition, we also present 
detailed results for the DomainNet benchmark, without however including 
results for the combination of GGA with the baseline algorithms, due to 
computational restrictions. We also include experiments for the ColoredMNIST and RotatedMNIST datasets and report the average results over 5 runs. 

When applying GGA to existing methods, the only difference regarding the baseline algorithm training is that ``Algorithm 1'' (i.e. GGA)  is applied instead of the method’s update rules for the duration of the annealing process (training steps $A_s$ to $A_e$). The total epochs and method hyperparameters remain the same throughout training.

\begin{table*}
\centering
\small
\renewcommand{\arraystretch}{1.1}
\caption{\small{Out-of-domain accuracies (\%) on {PACS}.}}
\begin{tabular}{lllll|c}
\toprule
\textbf{Algorithm} & \textbf{A} & \textbf{C} & \textbf{P} & \textbf{S} & \textbf{Avg} \\
\midrule

IRM$^\dagger$ \cite{arjovsky2019irm}            & 84.8\scriptsize{$\pm1.3$}      \normalsize{({\color{red}$-3.6$})}& 
76.4\scriptsize{$\pm1.1$}      \normalsize{({\color{green}$+0.7$})} & 
96.7\scriptsize{$\pm0.6$}      \normalsize{({\color{red}$-0.3$})}& 
76.1\scriptsize{$\pm1.0$}      \normalsize{({\color{red}$-2.8$})} & 
83.5 \normalsize{({\color{red}$-1.5$})}          \\

ERM$^\ddagger$ \cite{vapnik1998statistical} & 
85.7 \scriptsize$\pm0.6$ & 
77.1 \scriptsize$\pm0.8$ & 
97.4 \scriptsize$\pm0.4$ & 
76.6 \scriptsize$\pm0.7$ & 
84.2 \\

GroupDRO$^\ddagger$ \cite{Sagawa2020GroupDRO}    & 
83.5\scriptsize{$\pm0.9$}    \normalsize{({\color{green}$+2.8$})}   & 
79.1\scriptsize{$\pm0.6$}    \normalsize{({\color{green}$+2.1$})}   & 
96.7\scriptsize{$\pm0.7$}    \normalsize{({\color{green}$+1.0$})}   & 
78.3\scriptsize{$\pm2.0$}    \normalsize{({\color{green}$+0.6$})}   & 
84.4 \normalsize{({\color{green}$+1.6$})}           \\

MTL$^\ddagger$\cite{blanchard2021mtl_marginal_transfer_learning}   & 87.5\scriptsize{$\pm0.8$}     \normalsize{({\color{red}$-1.5$})} & 
77.1\scriptsize{$\pm0.5$}     \normalsize{({\color{green}$+5.8$})} & 
96.4\scriptsize{$\pm0.8$}     \normalsize{({\color{green}$+0.6$})}   & 
77.3\scriptsize{$\pm1.8$}     \normalsize{({\color{green}$+2.4$})} &  
84.6 \normalsize{({\color{green}$+1.8$})}          \\

Mixup$^\dagger$ \cite{xu2020interdomain_mixup_aaai}             & 86.1\scriptsize{$\pm0.5$}   \normalsize{({\color{green}$+1.4$})} & 
78.9\scriptsize{$\pm0.8$}   \normalsize{({\color{green}$+0.3$})} & 
97.6\scriptsize{$\pm0.1$}   \normalsize{({\color{red}$-0.1$})} & 
75.8\scriptsize{$\pm1.8$}   \normalsize{({\color{green}$+2.2$})} & 
84.6   \normalsize{({\color{green}$+1.0$})}        \\

MMD$^\ddagger$ \cite{li2018mmd}                  & 86.1\scriptsize{$\pm1.4$}     \normalsize{({\color{black}$\pm0.0$})} & 
79.4\scriptsize{$\pm0.9$}     \normalsize{({\color{red}$-2.6$})} & 
96.6\scriptsize{$\pm0.2$}     \normalsize{({\color{green}$+0.6$})} & 76.5\scriptsize{$\pm0.5$}     \normalsize{({\color{green}$+5.4$})} & 
84.7 \normalsize{({\color{green}$+0.8$})}         \\

VREx$^\ddagger$ \cite{krueger2020vrex} & 
86.0\scriptsize{$\pm1.6$}     \normalsize{({\color{red}$-1.5$})} & 79.1\scriptsize{$\pm0.6$}     \normalsize{({\color{green}$+2.8$})} & 96.9\scriptsize{$\pm0.5$}     \normalsize{({\color{red}$-0.9$})} & 77.7\scriptsize{$\pm1.7$}     \normalsize{({\color{green}$+2.8$})} & 
84.9 \normalsize{({\color{green}$+0.6$})}         \\

MLDG$^\dagger$ \cite{li2018learning}                & 85.5\scriptsize{$\pm1.4$}   \normalsize{({\color{green}$+2.3$})} &
80.1\scriptsize{$\pm1.7$}   \normalsize{({\color{green}$+0.4$})} & 97.4\scriptsize{$\pm0.3$}   \normalsize{({\color{green}$-0.4$})} &
76.6\scriptsize{$\pm1.1$}   \normalsize{({\color{green}$+2.8$})} & 
84.9 \normalsize{({\color{green}$+1.2$})}      \\

ARM$^\ddagger$ \cite{zhang2020arm}               & 86.8\scriptsize{$\pm0.6$}     \normalsize{({\color{red}$-2.5$})} & 
76.8\scriptsize{$\pm0.5$}     \normalsize{({\color{green}$+3.7$})} & 
97.4\scriptsize{$\pm0.3$}     \normalsize{({\color{red}$-0.5$})} & 
79.3\scriptsize{$\pm1.2$}     \normalsize{({\color{green}$+0.2$})} & 
85.1 \normalsize{({\color{green}$+0.2$})}           \\

Mixstyle$^\ddagger$ \cite{zhou2021mixstyle}   & 
86.8\scriptsize{$\pm0.5$} \normalsize{({\color{green}$+1.1$})}            & 79.0\scriptsize{$\pm1.4$} \normalsize{({\color{green}$+1.3$})}          & 96.6\scriptsize{$\pm0.1$} \normalsize{({\color{red}$-1.1$})}          & 78.5\scriptsize{$\pm2.3$} \normalsize{({\color{green}$+4.0$})}          & 
85.2 \normalsize{({\color{green}$+1.4$})}           \\

CORAL$^\dagger$ \cite{sun2016coral}             & 88.3\scriptsize{$\pm0.2$}      \normalsize{({\color{red}$-0.3$})}& 
80.0\scriptsize{$\pm0.5$}      \normalsize{({\color{green}$+0.9$})} & 
97.5\scriptsize{$\pm0.3$}      \normalsize{({\color{red}$-1.0$})} & 
78.8\scriptsize{$\pm1.3$}      \normalsize{({\color{green}$+1.5$})}& 
86.2    \normalsize{({\color{green}$+0.3$})}       \\

SagNet$^\dagger$ \cite{nam2019sagnet}           &             87.4\scriptsize{$\pm0.2$}    \normalsize{({\color{red}$-2.3$})} & 
80.7\scriptsize{$\pm0.5$}    \normalsize{({\color{green}$+1.0$})} & 97.1\scriptsize{$\pm0.1$}    \normalsize{({\color{red}$-0.9$})}& 
80.0\scriptsize{$\pm1.0$}    \normalsize{({\color{red}$-1.8$})}& 
86.3 \normalsize{({\color{red}$-1.0$})}         \\

\midrule

RSC$^\dagger$ \cite{huang2020rsc}               & 85.4\scriptsize{$\pm0.9$}      \normalsize{({\color{red}$-1.8$})}& 
79.7\scriptsize{$\pm0.5$}      \normalsize{({\color{green}$+2.9$})}& 
97.6\scriptsize{$\pm0.9$}      \normalsize{({\color{red}$-1.0$})}& 
78.2\scriptsize{$\pm1.0$}      \normalsize{({\color{green}$+0.3$})}& 
85.2 \normalsize{({\color{green}$+0.1$})}   \\

SAM $^\ddagger$\cite{foret2020sharpness}  & 
85.6\scriptsize$\pm2.1$       \normalsize{({\color{green}$+3.1$})} & 
80.9\scriptsize$\pm1.2$       \normalsize{({\color{red}$-0.5$})} & 
97.0\scriptsize$\pm0.4$       \normalsize{({\color{red}$-0.4$})} & 
79.6\scriptsize$\pm1.6$       \normalsize{({\color{red}$-0.8$})} & 
85.8 \normalsize{({\color{green}$+0.3$})}   \\

GSAM $^\ddagger$ \cite{zhuang2022surrogate}       &
86.9\scriptsize$\pm0.1$      \normalsize{({\color{red}$-1.9$})} & 
80.4\scriptsize$\pm0.2$      \normalsize{({\color{green}$+1.9$})} & 
97.5\scriptsize$\pm0.0$      \normalsize{({\color{red}$-1.5$})}& 
78.7\scriptsize$\pm0.8$      \normalsize{({\color{green}$+2.5$})}& 
85.9  \normalsize{({\color{green}$+0.6$})}  \\

SAGM $^\ddagger$ \cite{wang2023sharpness}       & 
87.4\scriptsize{$\pm0.2$}    \normalsize{({\color{green}$+1.2$})}& 
80.2\scriptsize{$\pm0.3$}    \normalsize{({\color{green}$+1.1$})}& 
98.0\scriptsize{$\pm0.2$}    \normalsize{({\color{red}$-1.0$})}& 
80.8\scriptsize{$\pm0.6$}    \normalsize{({\color{red}$-0.4$})}& 
86.6 \normalsize{({\color{green}$+0.2$})}   \\

\midrule
\textbf{GGA} (ours)                 & 
86.5\scriptsize{$\pm1.8$}           & 
81.2\scriptsize{$\pm3.0$}           & 
97.1\scriptsize{$\pm0.9$}           &  
80.8\scriptsize{$\pm0.9$}    &  
86.4  \\

\textbf{GGA-L} (ours)                 & 
88.0\scriptsize{$\pm1.0$}           & 
81.2\scriptsize{$\pm2.0$}           & 
97.1\scriptsize{$\pm0.3$}           &  
80.8\scriptsize{$\pm2.5$}    &  
86.5  \\

\bottomrule
\end{tabular}
\end{table*}

\begin{table*}[]
\centering
\small
\renewcommand{\arraystretch}{1.1}
\caption{\small{Out-of-domain accuracies (\%) on VLCS.}}
\begin{tabular}{lllll|c}
\toprule
\textbf{Algorithm} & \textbf{C} & \textbf{L} & \textbf{S} & \textbf{V} & \textbf{Avg} \\
\midrule

GroupDRO$^\ddagger$ \cite{Sagawa2020GroupDRO}    & 97.3\scriptsize{$\pm0.3$}    \normalsize{({\color{green}$+1.4$})}   & 
63.4\scriptsize{$\pm0.9$}    \normalsize{({\color{green}$+1.7$})}   & 
69.5\scriptsize{$\pm0.8$}    \normalsize{({\color{green}$+2.0$})}   & 
76.7\scriptsize{$\pm0.7$}    \normalsize{({\color{red}$-2.9$})}   & 
76.7 \normalsize{({\color{green}$+0.6$})}           \\

MLDG$^\dagger$ \cite{li2018learning}                & 97.4\scriptsize{$\pm0.2$}      \normalsize{({\color{green}$+1.6$})} &
65.2\scriptsize{$\pm0.7$}      \normalsize{({\color{green}$+0.4$})} & 71.0\scriptsize{$\pm1.4$}      \normalsize{({\color{green}$+0.6$})} &
75.3\scriptsize{$\pm1.0$}      \normalsize{({\color{green}$+2.7$})} & 
77.2 \normalsize{({\color{green}$+1.3$})}      \\

MTL$^\ddagger$\cite{blanchard2021mtl_marginal_transfer_learning}    & 97.8\scriptsize{$\pm0.4$}     \normalsize{({\color{green}$+0.3$})} & 
64.3\scriptsize{$\pm0.3$}     \normalsize{({\color{green}$+2.1$})} & 
71.5\scriptsize{$\pm0.7$}     \normalsize{({\color{red}$-0.3$})} & 
75.3\scriptsize{$\pm1.7$}     \normalsize{({\color{green}$+1.5$})} &  
77.2 \normalsize{({\color{green}$+0.9$})}          \\

ERM$^\ddagger$ \cite{vapnik1998statistical}  & 
98.0 \scriptsize$\pm0.3$ & 
64.7 \scriptsize$\pm1.2$ & 
71.4 \scriptsize$\pm1.2$ & 
75.2 \scriptsize$\pm1.6$ & 
77.3 \\

Mixup$^\dagger$ \cite{xu2020interdomain_mixup_aaai}             & 98.3\scriptsize{$\pm0.6$}     \normalsize{({\color{red}$-0.3$})} & 
64.8\scriptsize{$\pm1.0$}     \normalsize{({\color{red}$-1.9$})} & 
72.1\scriptsize{$\pm0.5$}     \normalsize{({\color{red}$-0.1$})} & 
74.3\scriptsize{$\pm0.8$}     \normalsize{({\color{green}$+4.0$})} & 
77.4  \normalsize{({\color{green}$+0.4$})}        \\

MMD$^\ddagger$ \cite{li2018mmd}                  & 97.7\scriptsize{$\pm0.1$}     \normalsize{({\color{green}$+1.2$})} & 
64.0\scriptsize{$\pm1.1$}     \normalsize{({\color{green}$+0.3$})} & 
72.8\scriptsize{$\pm0.2$}     \normalsize{({\color{red}$-0.7$})} & 
75.3\scriptsize{$\pm3.3$}     \normalsize{({\color{green}$+2.0$})} & 
77.5 \normalsize{({\color{green}$+0.6$})}         \\

ARM$^\ddagger$ \cite{zhang2020arm}               & 98.7\scriptsize{$\pm0.2$}     \normalsize{({\color{green}$+0.2$})}  & 
63.6\scriptsize{$\pm0.7$}     \normalsize{({\color{green}$+2.8$})}  & 
71.3\scriptsize{$\pm1.2$}     \normalsize{({\color{black}$\pm0.0$})}  & 
76.7\scriptsize{$\pm0.6$}     \normalsize{({\color{red}$-1.5$})}  & 
77.6 \normalsize{({\color{green}$+1.4$})}           \\

SagNet$^\dagger$ \cite{nam2019sagnet}           &             97.9\scriptsize{$\pm0.4$}     \normalsize{({\color{red}$-0.2$})} & 
64.5\scriptsize{$\pm0.5$}     \normalsize{({\color{green}$+1.7$})} & 71.4\scriptsize{$\pm1.3$}     \normalsize{({\color{green}$+0.8$})}& 
77.5\scriptsize{$\pm0.5$}     \normalsize{({\color{green}$+1.4$})}& 
77.8 \normalsize{({\color{green}$+0.9$})}         \\

Mixstyle$^\ddagger$ \cite{zhou2021mixstyle}   & 
98.6\scriptsize{$\pm0.3$} \normalsize{({\color{red}$-0.1$})}            & 64.5\scriptsize{$\pm1.1$} \normalsize{({\color{green}$+1.9$})}          & 72.6\scriptsize{$\pm0.5$} \normalsize{({\color{green}$+0.4$})}          & 75.7\scriptsize{$\pm1.7$} \normalsize{({\color{green}$+0.4$})}          & 
77.9 \normalsize{({\color{green}$+0.6$})}           \\

VREx$^\ddagger$ \cite{krueger2020vrex}         & 
98.4\scriptsize{$\pm0.3$}      \normalsize{({\color{red}$-0.9$})} & 64.4\scriptsize{$\pm1.4$}      \normalsize{({\color{green}$+1.9$})} & 74.1\scriptsize{$\pm0.4$}      \normalsize{({\color{green}$-1.7$})} & 76.2\scriptsize{$\pm1.3$}      \normalsize{({\color{green}$+1.0$})} & 
78.3 \normalsize{({\color{green}$+0.1$})}         \\

IRM$^\dagger$ \cite{arjovsky2019irm}            & 98.6\scriptsize{$\pm0.1$}      \normalsize{({\color{red}$-0.3$})}& 
64.9\scriptsize{$\pm0.9$}      \normalsize{({\color{red}$-3.5$})}& 
73.4\scriptsize{$\pm0.6$}      \normalsize{({\color{green}$+1.7$})}& 
77.3\scriptsize{$\pm0.9$}      \normalsize{({\color{red}$-1.5$})} & 
78.6 \normalsize{({\color{red}$-0.9$})}          \\

CORAL$^\dagger$ \cite{sun2016coral}             & 98.3\scriptsize{$\pm0.3$}      \normalsize{({\color{green}$+0.9$})}& 
66.1\scriptsize{$\pm0.6$}      \normalsize{({\color{green}$+1.8$})} & 
73.4\scriptsize{$\pm0.3$}      \normalsize{({\color{red}$-1.8$})} & 
77.5\scriptsize{$\pm1.0$}      \normalsize{({\color{red}$-2.1$})}& 
78.8    \normalsize{({\color{red}$-0.4$})}       \\

\midrule

RSC$^\dagger$ \cite{huang2020rsc}               & 
97.9\scriptsize{$\pm0.1$}    \normalsize{({\color{green}$+0.6$})}& 62.5\scriptsize{$\pm0.7$}    \normalsize{({\color{green}$+0.3$})}& 
72.3\scriptsize{$\pm1.2$}    \normalsize{({\color{green}$+0.4$})}& 
75.6\scriptsize{$\pm0.8$}    \normalsize{({\color{red}$-0.8$})}& 
77.1 \normalsize{({\color{green}$+0.2$})}   \\


GSAM $^\ddagger$ \cite{zhuang2022surrogate}       & 
98.7\scriptsize$\pm0.3$   \normalsize{({\color{green}$+0.9$})} & 
64.9\scriptsize$\pm0.2$   \normalsize{({\color{green}$+0.7$})} & 
74.3\scriptsize$\pm0.0$   \normalsize{({\color{red}$-1.9$})}& 
78.5\scriptsize$\pm0.8$   \normalsize{({\color{green}$+0.3$})}& 
79.1  \normalsize{({\color{black}$\pm0.0$})}  \\

SAM $^\ddagger$\cite{foret2020sharpness}  & 
99.1\scriptsize$\pm0.2$     \normalsize{({\color{red}$-0.2$})} & 
65.0\scriptsize$\pm1.0$     \normalsize{({\color{green}$+0.2$})} & 
73.7\scriptsize$\pm1.0$     \normalsize{({\color{red}$-0.4$})} & 
79.8\scriptsize$\pm0.1$     \normalsize{({\color{green}$+1.1$})} & 
79.4 \normalsize{({\color{green}$+0.2$})}   \\

SAGM $^\ddagger$ \cite{wang2023sharpness}       & 
99.0\scriptsize{$\pm0.2$}  \normalsize{({\color{red}$-0.2$})}& 
65.2\scriptsize{$\pm0.4$}  \normalsize{({\color{green}$+0.4$})}& 
75.1\scriptsize{$\pm0.3$}  \normalsize{({\color{green}$+0.2$})}& 
80.7\scriptsize{$\pm0.8$}  \normalsize{({\color{red}$-0.4$})}& 
80.0 \normalsize{({\color{black}$\pm0.0$})}   \\

\midrule
\textbf{GGA} (ours)                 & 
98.4\scriptsize{$\pm0.2$}           & 
65.4\scriptsize{$\pm0.1$}           & 
73.8\scriptsize{$\pm1.6$}           &  
77.4\scriptsize{$\pm1.9$}    &  
78.7  \\

\textbf{GGA-L} (ours)                 & 
98.9\scriptsize{$\pm0.4$}           & 
66.5\scriptsize{$\pm0.3$}           & 
70.0\scriptsize{$\pm2.0$}           &  
78.1\scriptsize{$\pm1.1$}    &  
78.4  \\
\bottomrule
\end{tabular}
\end{table*}

\begin{table*}[]
\centering
\small
\renewcommand{\arraystretch}{1.1}
\caption{\small{Out-of-domain accuracies (\%) on OfficeHome.}}
\begin{tabular}{lllll|c}
\toprule
\textbf{Algorithm} & \textbf{A} & \textbf{C} & \textbf{P} & \textbf{R} & \textbf{Avg} \\
\midrule
Mixstyle$^\ddagger$ \cite{zhou2021mixstyle}   & 
51.1\scriptsize{$\pm0.3$} \normalsize{({\color{green}$+0.3$})}            & 53.2\scriptsize{$\pm0.4$} \normalsize{({\color{green}$+0.7$})}          & 68.2\scriptsize{$\pm0.7$} \normalsize{({\color{green}$+0.4$})}          & 69.2\scriptsize{$\pm0.6$} \normalsize{({\color{green}$+0.6$})}          & 
60.4 \normalsize{({\color{green}$+0.5$})}           \\

IRM$^\dagger$ \cite{arjovsky2019irm}            & 58.9\scriptsize{$\pm2.3$}       \normalsize{({\color{red}$-3.5$})}& 
52.2\scriptsize{$\pm1.6$}       \normalsize{({\color{red}$-2.1$})} & 
72.1\scriptsize{$\pm2.9$}       \normalsize{({\color{red}$-1.7$})}& 
74.0\scriptsize{$\pm2.5$}       \normalsize{({\color{red}$-1.0$})} & 
64.3 \normalsize{({\color{red}$-2.1$})}          \\

ARM$^\ddagger$ \cite{zhang2020arm}               & 58.9\scriptsize{$\pm0.8$}      \normalsize{({\color{red}$-0.7$})} & 
51.0\scriptsize{$\pm0.5$}      \normalsize{({\color{green}$+1.8$})} & 
74.1\scriptsize{$\pm0.1$}      \normalsize{({\color{green}$+2.2$})} & 
75.2\scriptsize{$\pm0.3$}      \normalsize{({\color{green}$+1.6$})} & 
64.8 \normalsize{({\color{green}$+1.2$})}           \\

GroupDRO$^\ddagger$ \cite{Sagawa2020GroupDRO}    & 
60.4\scriptsize{$\pm0.7$}    \normalsize{({\color{green}$+0.4$})}   & 
52.7\scriptsize{$\pm1.0$}    \normalsize{({\color{red}$-0.6$})}   & 
75.0\scriptsize{$\pm0.7$}    \normalsize{({\color{red}$-1.1$})}   & 
76.0\scriptsize{$\pm0.7$}    \normalsize{({\color{red}$-0.3$})}   & 
66.0 \normalsize{({\color{red}$-0.4$})}           \\

MMD$^\ddagger$ \cite{li2018mmd}                  & 60.4\scriptsize{$\pm0.2$}     \normalsize{({\color{red}$-0.6$})} & 
53.3\scriptsize{$\pm0.3$}     \normalsize{({\color{green}$+2.4$})} & 
74.3\scriptsize{$\pm0.1$}     \normalsize{({\color{green}$+1.9$})} & 
77.4\scriptsize{$\pm0.6$}     \normalsize{({\color{red}$-1.3$})} & 
66.4 \normalsize{({\color{green}$+0.3$})}         \\

MTL$^\ddagger$\cite{blanchard2021mtl_marginal_transfer_learning}    & 61.5\scriptsize{$\pm0.7$}       \normalsize{({\color{red}$-0.7$})} & 
52.4\scriptsize{$\pm0.6$}       \normalsize{({\color{green}$+0.6$})} & 
74.9\scriptsize{$\pm0.4$}       \normalsize{({\color{red}$-0.7$})} & 
76.8\scriptsize{$\pm0.4$}       \normalsize{({\color{red}$-0.6$})}&  
66.4 \normalsize{({\color{red}$-0.3$})}          \\

VREx$^\ddagger$ \cite{krueger2020vrex}         & 
60.7\scriptsize{$\pm0.9$}        \normalsize{({\color{green}$+1.8$})} & 53.0\scriptsize{$\pm0.9$}        \normalsize{({\color{green}$+2.0$})} & 75.3\scriptsize{$\pm0.1$}        \normalsize{({\color{green}$+1.1$})} & 76.6\scriptsize{$\pm0.5$}        \normalsize{({\color{green}$+1.1$})} & 
66.4 \normalsize{({\color{green}$+1.5$})}         \\

ERM$^\ddagger$ \cite{vapnik1998statistical} & 
63.1 \scriptsize$\pm0.3$ & 
51.9 \scriptsize$\pm0.4$ & 
77.2 \scriptsize$\pm0.5$ & 
78.1 \scriptsize$\pm0.2$ & 67.6 \\

MLDG$^\dagger$ \cite{li2018learning}                & 61.5\scriptsize{$\pm0.9$}       \normalsize{({\color{green}$+2.4$})} &
53.2\scriptsize{$\pm0.6$}       \normalsize{({\color{green}$+0.2$})} & 75.0\scriptsize{$\pm1.2$}       \normalsize{({\color{green}$+1.8$})} &
77.5\scriptsize{$\pm0.4$}       \normalsize{({\color{green}$+0.6$})} & 
66.8 \normalsize{({\color{green}$+1.2$})}      \\

Mixup$^\dagger$ \cite{xu2020interdomain_mixup_aaai}             & 62.4\scriptsize{$\pm0.8$}       \normalsize{({\color{green}$+0.6$})} & 
54.8\scriptsize{$\pm0.6$}       \normalsize{({\color{red}$-0.2$})} & 
76.9\scriptsize{$\pm0.3$}       \normalsize{({\color{green}$+0.2$})} & 
78.3\scriptsize{$\pm0.2$}       \normalsize{({\color{green}$+0.4$})} & 
68.1   \normalsize{({\color{green}$+0.1$})}        \\

SagNet$^\dagger$ \cite{nam2019sagnet}           &             63.4\scriptsize{$\pm0.2$}       \normalsize{({\color{green}$+1.0$})} & 
54.8\scriptsize{$\pm0.4$}       \normalsize{({\color{red}$-1.9$})} & 75.8\scriptsize{$\pm0.4$}       \normalsize{({\color{green}$+1.4$})}& 
78.3\scriptsize{$\pm0.3$}       \normalsize{({\color{green}$+0.7$})}& 
68.1 \normalsize{({\color{green}$+0.3$})}         \\

CORAL$^\dagger$ \cite{sun2016coral}             & 
65.3\scriptsize{$\pm0.3$}       \normalsize{({\color{green}$+0.3$})}& 
54.4\scriptsize{$\pm0.6$}       \normalsize{({\color{green}$+0.2$})} & 
76.5\scriptsize{$\pm0.3$}       \normalsize{({\color{red}$-0.7$})} & 
78.4\scriptsize{$\pm1.0$}       \normalsize{({\color{green}$+1.0$})}& 
68.7    \normalsize{({\color{green}$+0.2$})}       \\

\midrule

RSC$^\dagger$ \cite{huang2020rsc}               & 
60.7\scriptsize{$\pm1.4$}      \normalsize{({\color{red}$-1.1$})}& 
51.4\scriptsize{$\pm0.3$}      \normalsize{($+0.0$)}& 
74.8\scriptsize{$\pm1.1$}      \normalsize{({\color{green}$+0.6$})}& 
75.1\scriptsize{$\pm1.3$}      \normalsize{({\color{green}$+0.5$})}& 
65.5 \normalsize{($+0.0$)}  \\


GSAM $^\ddagger$ \cite{zhuang2022surrogate}  & 
64.9\scriptsize$\pm0.1$     \normalsize{({\color{red}$-3.4$})} & 
55.2\scriptsize$\pm0.2$     \normalsize{({\color{green}$+1.5$})} & 
77.8\scriptsize$\pm0.0$     \normalsize{({\color{green}$+0.3$})}& 
79.2\scriptsize$\pm0.2$     \normalsize{({\color{red}$-1.0$})}& 
69.3  \normalsize{({\color{red}$-0.7$})}  \\

SAM $^\ddagger$\cite{foret2020sharpness}  & 
64.5\scriptsize$\pm0.3$     \normalsize{({\color{red}$-0.8$})} & 
56.5\scriptsize$\pm0.2$     \normalsize{({\color{red}$-0.5$})} & 
77.4\scriptsize$\pm0.1$     \normalsize{({\color{red}$-2.1$})} & 
79.8\scriptsize$\pm0.4$     \normalsize{({\color{red}$-0.7$})} & 
69.6 \normalsize{({\color{red}$-1.0$})}   \\

SAGM $^\ddagger$ \cite{wang2023sharpness}       & 
65.4\scriptsize{$\pm0.4$}    \normalsize{({\color{red}$-0.9$})}& 
57.0\scriptsize{$\pm0.3$}    \normalsize{({\color{red}$-0.8$})}& 
78.0\scriptsize{$\pm0.3$}    \normalsize{({\color{green}$+0.4$})}& 
80.0\scriptsize{$\pm0.2$}    \normalsize{({\color{red}$-1.1$})}& 
70.1 \normalsize{({\color{red}$-0.6$})}   \\

\midrule
\textbf{GGA} (ours)                 & 
61.7\scriptsize{$\pm0.1$}           & 
52.5\scriptsize{$\pm0.5$}           & 
77.1\scriptsize{$\pm1.3$}           &  
77.0\scriptsize{$\pm0.1$}    &  
67.0  \\

\textbf{GGA-L} (ours)                 & 
59.7\scriptsize{$\pm0.2$}           & 
53.8\scriptsize{$\pm0.5$}           & 
75.3\scriptsize{$\pm0.8$}           &  
77.1\scriptsize{$\pm0.1$}    &  
66.5  \\

\bottomrule
\end{tabular}
\end{table*}

\begin{table*}[]
\centering
\small
\renewcommand{\arraystretch}{1.1}
\caption{\small{Out-of-domain accuracies (\%) on TerraIncognita.}}
\begin{tabular}{lllll|c}
\toprule
\textbf{Algorithm} & \textbf{L100} & \textbf{L38} & \textbf{L43} & \textbf{L46} & \textbf{Avg} \\
\midrule

MMD$^\ddagger$ \cite{li2018mmd}                  & 41.9\scriptsize{$\pm3.0$}     \normalsize{({\color{green}$+2.1$})} & 
34.8\scriptsize{$\pm1.0$}     \normalsize{({\color{green}$+9.1$})} & 
57.0\scriptsize{$\pm1.9$}     \normalsize{({\color{green}$+3.0$})} & 35.2\scriptsize{$\pm1.8$}     \normalsize{({\color{green}$+4.0$})} & 
42.2 \normalsize{({\color{green}$+5.1$})}         \\

GroupDRO$^\ddagger$ \cite{Sagawa2020GroupDRO}    & 
41.2\scriptsize{$\pm0.7$}    \normalsize{({\color{green}$+8.9$})}   & 
38.6\scriptsize{$\pm2.1$}    \normalsize{({\color{green}$+2.9$})}   & 
56.7\scriptsize{$\pm0.9$}    \normalsize{({\color{green}$+2.4$})}   & 
36.4\scriptsize{$\pm2.1$}    \normalsize{({\color{green}$+5.7$})}   & 
43.2 \normalsize{({\color{green}$+1.7$})}           \\

Mixstyle$^\ddagger$ \cite{zhou2021mixstyle}   & 
54.3\scriptsize{$\pm1.1$} \normalsize{({\color{green}$+6.8$})}            & 34.1\scriptsize{$\pm1.1$} \normalsize{({\color{green}$+8.8$})}          & 55.9\scriptsize{$\pm1.1$} \normalsize{({\color{green}$+1.6$})}          & 31.7\scriptsize{$\pm2.1$} \normalsize{({\color{red}$-1.1$})}          & 
44.0 \normalsize{({\color{green}$+4.0$})}           \\

ARM$^\ddagger$ \cite{zhang2020arm}               & 49.3\scriptsize{$\pm0.7$}     \normalsize{({\color{green}$+3.9$})} & 
38.3\scriptsize{$\pm0.7$}     \normalsize{({\color{green}$+7.2$})} & 
55.8\scriptsize{$\pm0.8$}     \normalsize{({\color{red}$-0.5$})} & 
38.7\scriptsize{$\pm1.3$}     \normalsize{({\color{red}$-0.8$})} & 
45.5 \normalsize{({\color{green}$+2.2$})}           \\

MTL$^\ddagger$\cite{blanchard2021mtl_marginal_transfer_learning}    & 49.3\scriptsize{$\pm1.2$}      \normalsize{({\color{green}$+2.2$})} & 
39.6\scriptsize{$\pm6.3$}      \normalsize{({\color{red}$-2.7$})} & 
55.6\scriptsize{$\pm1.1$}      \normalsize{({\color{green}$+3.3$})} & 
37.8\scriptsize{$\pm0.8$}      \normalsize{({\color{green}$+0.7$})}&  
45.6 \normalsize{({\color{green}$+0.9$})}          \\

ERM$^\ddagger$  \cite{vapnik1998statistical}   & 
49.8 \scriptsize$\pm4.4$ & 
42.1 \scriptsize$\pm1.4$ & 
56.9 \scriptsize$\pm1.8$ & 
35.7 \scriptsize$\pm3.9$ & 
46.1 \\

VREx$^\ddagger$ \cite{krueger2020vrex}         & 
48.2\scriptsize{$\pm4.3$}       \normalsize{({\color{green}$+2.3$})}& 41.7\scriptsize{$\pm1.3$}       \normalsize{({\color{red}$-4.8$})}& 56.8\scriptsize{$\pm0.8$}       \normalsize{({\color{green}$+2.1$})}& 38.7\scriptsize{$\pm3.1$}       \normalsize{({\color{red}$-0.2$})} & 
46.4 \normalsize{({\color{green}$+0.1$})}         \\

IRM$^\dagger$ \cite{arjovsky2019irm}            & 54.6\scriptsize{$\pm1.3$}      \normalsize{({\color{red}$-4.3$})}& 
39.8\scriptsize{$\pm1.9$}      \normalsize{({\color{red}$-3.4$})} & 
56.2\scriptsize{$\pm1.8$}      \normalsize{({\color{red}$-3.8$})}& 
39.6\scriptsize{$\pm0.8$}      \normalsize{({\color{red}$-4.1$})} & 
47.6 \normalsize{({\color{red}$-3.9$})}          \\

CORAL$^\dagger$ \cite{sun2016coral}             & 
51.6\scriptsize{$\pm2.4$}      \normalsize{({\color{green}$+3.1$})}& 
42.2\scriptsize{$\pm1.0$}      \normalsize{({\color{red}$-1.2$})} & 
57.0\scriptsize{$\pm1.0$}      \normalsize{({\color{green}$+1.1$})} & 
39.8\scriptsize{$\pm2.9$}      \normalsize{({\color{red}$-1.8$})}& 
47.6    \normalsize{({\color{green}$+0.3$})}       \\

MLDG$^\dagger$ \cite{li2018learning}                & 54.2\scriptsize{$\pm3.0$}       \normalsize{({\color{red}$-3.4$})} &
44.3\scriptsize{$\pm1.1$}       \normalsize{({\color{green}$+2.3$})} & 55.6\scriptsize{$\pm0.3$}       \normalsize{({\color{green}$+1.2$})} &
36.9\scriptsize{$\pm2.2$}       \normalsize{({\color{green}$+0.4$})} & 
47.8 \normalsize{({\color{green}$+0.1$})}      \\

Mixup$^\dagger$ \cite{xu2020interdomain_mixup_aaai}             & 59.6\scriptsize{$\pm2.0$}       \normalsize{({\color{green}$+1.6$})} & 
42.2\scriptsize{$\pm1.4$}       \normalsize{({\color{green}$+3.5$})} & 
55.9\scriptsize{$\pm0.8$}       \normalsize{({\color{green}$+0.5$})} & 
33.9\scriptsize{$\pm1.4$}       \normalsize{({\color{red}$-3.5$})} & 
47.9   \normalsize{({\color{green}$+0.5$})}        \\

SagNet$^\dagger$ \cite{nam2019sagnet}           &             53.0\scriptsize{$\pm2.0$}       \normalsize{({\color{green}$+2.3$})} & 
43.0\scriptsize{$\pm1.4$}       \normalsize{({\color{green}$+0.2$})} & 57.9\scriptsize{$\pm0.8$}       \normalsize{({\color{red}$-2.6$})}& 
40.4\scriptsize{$\pm1.4$}       \normalsize{({\color{green}$+2.9$})}& 
48.6 \normalsize{({\color{green}$+0.4$})}         \\

\midrule

SAM $^\ddagger$\cite{foret2020sharpness}  & 
46.3\scriptsize$\pm1.0$          \normalsize{({\color{green}$+6.6$})} & 
38.4\scriptsize$\pm2.4$          \normalsize{({\color{green}$+1.2$})} & 
54.0\scriptsize$\pm1.0$          \normalsize{({\color{green}$+1.2$})} & 
34.5\scriptsize$\pm0.8$          \normalsize{({\color{green}$+2.7$})} & 
43.3 \normalsize{({\color{green}$+2.9$})}   \\

RSC$^\dagger$ \cite{huang2020rsc}               & 
50.2\scriptsize{$\pm2.2$}      \normalsize{({\color{red}$-0.8$})}& 
39.2\scriptsize{$\pm1.4$}      \normalsize{({\color{green}$+1.0$})}& 
56.3\scriptsize{$\pm1.4$}      \normalsize{({\color{green}$+0.8$})}&
40.8\scriptsize{$\pm0.6$}      \normalsize{({\color{green}$+0.2$})}& 
46.6 \normalsize{({\color{green}$+0.2$})}   \\

GSAM $^\ddagger$ \cite{zhuang2022surrogate} & 
50.8\scriptsize$\pm0.1$        \normalsize{({\color{green}$+1.4$})} & 
39.3\scriptsize$\pm0.2$        \normalsize{({\color{green}$+0.9$})} & 
59.6\scriptsize$\pm0.0$        \normalsize{({\color{red}$-2.9$})}& 
38.2\scriptsize$\pm0.8$        \normalsize{({\color{red}$-4.0$})}& 
47.0  \normalsize{({\color{red}$-1.0$})}  \\

SAGM $^\ddagger$ \cite{wang2023sharpness}       & 
54.8\scriptsize{$\pm1.3$}   \normalsize{($\pm0.0$)}& 
41.4\scriptsize{$\pm0.8$}   \normalsize{({\color{green}$+6.3$})}& 
57.7\scriptsize{$\pm0.6$}   \normalsize{({\color{red}$-1.1$})}& 
41.3\scriptsize{$\pm0.4$}   \normalsize{({\color{red}$-5.5$})}& 
48.8 \normalsize{({\color{red}$-0.1$})}   \\

\midrule
\textbf{GGA} (ours)                 & 
50.9\scriptsize{$\pm2.2$}           & 
42.5\scriptsize{$\pm1.0$}           & 
59.7\scriptsize{$\pm1.4$}           &  
41.5\scriptsize{$\pm3.5$}    &  
48.5  \\

\textbf{GGA-L} (ours)                 & 
57.2\scriptsize{$\pm5.2$}           & 
45.1\scriptsize{$\pm1.0$}           & 
56.4\scriptsize{$\pm1.4$}           &  
44.5\scriptsize{$\pm3.5$}    &  
49.8  \\

\bottomrule
\end{tabular}
\end{table*}

\begin{table*}[]
\centering
\small
\renewcommand{\arraystretch}{1.1}
\caption{\small{Out-of-domain accuracies (\%) on {DomainNet}.}}
\begin{tabular}{lllllll|c}
\toprule
\textbf{Algorithm} & \textbf{clip} & \textbf{info} & \textbf{paint} & \textbf{quick} & \textbf{real} & \textbf{sketch} & \textbf{Avg} \\
\midrule

MMD$^\dagger$ \cite{li2018mmd}   & 
32.1 \scriptsize$\pm13.3$ & 
11.0 \scriptsize$\pm4.6$ & 
26.8 \scriptsize$\pm11.3$ & 
8.7 \scriptsize$\pm2.1$ & 
32.7 \scriptsize$\pm13.8$ & 
28.9 \scriptsize$\pm11.9$ & 23.4 \\

GroupDRO$^\dagger$ \cite{Sagawa2020GroupDRO} & 
47.2 \scriptsize$\pm0.5$ & 
17.5 \scriptsize$\pm0.4$ & 
33.8 \scriptsize$\pm0.5$ & 
9.3 \scriptsize$\pm0.3$ & 
51.6 \scriptsize$\pm0.4$ & 
40.1 \scriptsize$\pm0.6$ & 33.3 \\

VREx$^\dagger$ \cite{krueger2020vrex} & 
47.3 \scriptsize$\pm3.5$ & 
16.0 \scriptsize$\pm1.5$ & 
35.8 \scriptsize$\pm4.6$ & 
10.9 \scriptsize$\pm0.3$ & 
49.6 \scriptsize$\pm4.9$ & 
42.0 \scriptsize$\pm3.0$ & 33.6 \\

IRM$^\dagger$ \cite{arjovsky2019irm}  & 
48.5 \scriptsize$\pm2.8$ & 
15.0 \scriptsize$\pm1.5$ & 
38.3 \scriptsize$\pm4.3$ & 
10.9 \scriptsize$\pm0.5$ & 
48.2 \scriptsize$\pm5.2$ & 
42.3 \scriptsize$\pm3.1$ & 33.9 \\

Mixstyle$^\ddagger$ \cite{zhou2021mixstyle} & 
51.9 \scriptsize$\pm0.4$ & 
13.3 \scriptsize$\pm0.2$ & 
37.0 \scriptsize$\pm0.5$ & 
12.3 \scriptsize$\pm0.1$ & 
46.1 \scriptsize$\pm0.3$ & 
43.4 \scriptsize$\pm0.4$ & 34.0 \\

ARM$^\dagger$ \cite{zhang2020arm} & 
49.7 \scriptsize$\pm0.3$ & 
16.3 \scriptsize$\pm0.5$ & 
40.9 \scriptsize$\pm1.1$ & 
9.4 \scriptsize$\pm0.1$ & 
53.4 \scriptsize$\pm0.4$ & 
43.5 \scriptsize$\pm0.4$ & 35.5 \\

Mixup$^\ddagger$ \cite{xu2020interdomain_mixup_aaai} & 
55.7\scriptsize$\pm0.3$ & 
18.5\scriptsize$\pm0.5$ & 
44.3\scriptsize$\pm0.5$ & 
12.5\scriptsize$\pm0.4$ & 
55.8\scriptsize$\pm0.3$ & 
48.2\scriptsize$\pm0.5$ & 39.2 \\

SagNet$^\dagger$ \cite{nam2019sagnet} & 
57.7 \scriptsize$\pm0.3$ & 
19.0 \scriptsize$\pm0.2$ & 
45.3 \scriptsize$\pm0.3$ & 
12.7 \scriptsize$\pm0.5$ & 
58.1 \scriptsize$\pm0.5$ & 
48.8 \scriptsize$\pm0.2$ & 40.3 \\

MTL$^\dagger$ \cite{blanchard2021mtl_marginal_transfer_learning}& 
57.9 \scriptsize$\pm0.5$ & 
18.5 \scriptsize$\pm0.4$ & 
46.0 \scriptsize$\pm0.1$ & 
12.5 \scriptsize$\pm0.1$ & 
59.5 \scriptsize$\pm0.3$ & 
49.2 \scriptsize$\pm0.1$ & 40.6 \\

MLDG$^\dagger$ \cite{li2018learning}& 
59.1 \scriptsize$\pm0.2$ & 
19.1 \scriptsize$\pm0.3$ & 
45.8 \scriptsize$\pm0.7$ & 
13.4 \scriptsize$\pm0.3$ & 
59.6 \scriptsize$\pm0.2$ & 
50.2 \scriptsize$\pm0.4$ & 41.2 \\

CORAL$^\dagger$ \cite{sun2016coral}& 
59.2 \scriptsize$\pm0.1$ & 
19.7 \scriptsize$\pm0.2$ & 
46.6 \scriptsize$\pm0.3$ & 
13.4 \scriptsize$\pm0.4$ &
59.8 \scriptsize$\pm0.2$ & 
50.1 \scriptsize$\pm0.6$ & 41.5 \\

ERM$^\ddagger$ \cite{vapnik1998statistical} & 
63.0 \scriptsize$\pm0.2$ & 
21.2 \scriptsize$\pm0.2$ & 
50.1 \scriptsize$\pm0.4$ & 
13.9 \scriptsize$\pm0.5$ & 
63.7 \scriptsize$\pm0.2$ &
52.0 \scriptsize$\pm0.5$ & 43.8 \\

\midrule 

RSC$^\dagger$ \cite{huang2020rsc}               & 
55.0\scriptsize{$\pm1.2$} & 
18.3\scriptsize{$\pm0.5$} & 
44.4\scriptsize{$\pm0.6$} &
12.2\scriptsize{$\pm0.2$} & 
55.7\scriptsize{$\pm0.7$} &
47.8\scriptsize{$\pm0.9$} &
38.9 \\

SAM$^\ddagger$ \cite{foret2020sharpness}  & 
64.5\scriptsize$\pm0.3$  & 
20.7\scriptsize$\pm0.2$  & 
50.2\scriptsize$\pm0.1$  & 
15.1\scriptsize$\pm0.3$  & 
62.6\scriptsize$\pm0.2$  & 
52.7\scriptsize$\pm0.3$  &  44.3   \\

GSAM $^\ddagger$ \cite{zhuang2022surrogate} & 
64.2\scriptsize$\pm0.3$  & 
20.8\scriptsize$\pm0.2$  & 
50.9\scriptsize$\pm0.0$  & 
14.4\scriptsize$\pm0.8$  & 
63.5\scriptsize$\pm0.2$  &
53.9\scriptsize$\pm0.2$  &  44.6   \\

SAGM $^\ddagger$ \cite{wang2023sharpness}   & 
64.9\scriptsize{$\pm0.2$}  & 
21.1\scriptsize{$\pm0.3$}  & 
51.5\scriptsize{$\pm0.2$}  & 
14.8\scriptsize{$\pm0.2$}  & 
64.1\scriptsize{$\pm0.2$}  & 
53.6\scriptsize{$\pm0.2$}  & 45.0 \\

\midrule

\textbf{GGA} (ours) & 
63.7\scriptsize{$\pm0.2$} & 
21.3\scriptsize{$\pm0.3$} & 
50.4\scriptsize{$\pm0.1$} & 
14.1\scriptsize{$\pm0.2$} & 
63.8\scriptsize{$\pm0.4$} & 
53.5\scriptsize{$\pm0.3$} & 44.4 \\

\textbf{GGA-L} (ours) & 
63.2\scriptsize{$\pm0.2$} & 
21.0\scriptsize{$\pm0.3$} & 
49.5\scriptsize{$\pm0.1$} & 
13.8\scriptsize{$\pm0.2$} & 
64.1\scriptsize{$\pm0.4$} & 
53.6\scriptsize{$\pm0.3$} & 44.5 \\

\bottomrule
\end{tabular}
\end{table*}

\begin{table*}
	\centering
	\small
	\renewcommand{\arraystretch}{1.1}
	\caption{\small{Out-of-domain accuracies (\%) on ColoredMNIST (left) and RotatedMNIST (right). All baseline results were reproduced.}}
	\begin{tabular}{llll|c|llllll|c}
		\toprule
		\textbf{Algorithm} & \textbf{0.1} & \textbf{0.2} & \textbf{0.9} & \textbf{Avg} 
		&\textbf{0} &\textbf{15} &\textbf{30} &
		\textbf{45} &\textbf{60} &
		\textbf{75} &\textbf{Avg} \\
		\midrule
		
		IRM$\ddagger$ \cite{arjovsky2019irm} & 
		56.8\scriptsize$\pm4.5$ & 63.5\scriptsize$\pm2.7$ & 
		10.2\scriptsize$\pm0.2$ & 
		43.5 &
		95.5\scriptsize{$\pm0.4$} & 98.7\scriptsize{$\pm0.2$} & 
		98.7\scriptsize{$\pm0.1$} & 98.5\scriptsize{$\pm0.3$} & 
		98.7\scriptsize{$\pm0.1$} & 96.1\scriptsize{$\pm0.1$} & 97.7\\ 
		
		MLDG \cite{li2018learning} & 
		71.5\scriptsize$\pm0.6$ & 73.0\scriptsize$\pm0.1$ & 
		10.1\scriptsize$\pm0.2$ & 
		51.5 &
		94.7\scriptsize{$\pm0.7$} & 98.8\scriptsize{$\pm0.1$} & 
		98.8\scriptsize{$\pm0.1$} & 98.8\scriptsize{$\pm0.1$} & 
		98.7\scriptsize{$\pm0.1$} & 95.9\scriptsize{$\pm0.4$} & 97.6\\ 
		
		MTL \cite{blanchard2021mtl_marginal_transfer_learning} & 
		71.3\scriptsize$\pm0.6$ & 72.9\scriptsize$\pm0.2$ & 
		10.2\scriptsize$\pm0.1$ & 
		51.5 &
		94.6\scriptsize{$\pm1.1$} & 98.6\scriptsize{$\pm0.2$} & 
		98.8\scriptsize{$\pm0.1$} & 98.7\scriptsize{$\pm0.1$} & 
		98.7\scriptsize{$\pm0.3$} & 95.3\scriptsize{$\pm0.7$} & 97.4\\ 
		
		Mixup \cite{xu2020interdomain_mixup_aaai} & 
		71.5\scriptsize$\pm0.8$ & 73.2\scriptsize$\pm0.3$ & 
		10.2\scriptsize$\pm0.2$ & 
		51.6 &
		94.9\scriptsize{$\pm0.5$} & 98.8\scriptsize{$\pm0.1$} & 
		98.8\scriptsize{$\pm0.2$} & 98.8\scriptsize{$\pm0.1$} & 
		98.8\scriptsize{$\pm0.1$} & 95.7\scriptsize{$\pm0.5$} & 97.6\\
		
		SagNet \cite{nam2019sagnet}  & 
		72.0\scriptsize$\pm0.5$ & 72.8\scriptsize$\pm0.5$ & 
		9.9\scriptsize$\pm0.3$ & 
		51.6 &
		95.5\scriptsize{$\pm0.3$} & 98.9\scriptsize{$\pm0.1$} & 
		99.0\scriptsize{$\pm0.1$} & 98.8\scriptsize{$\pm0.2$} & 
		98.8\scriptsize{$\pm0.1$} & 95.9\scriptsize{$\pm0.4$} & 97.8\\
		
		ERM \cite{vapnik1998statistical} & 
		71.8\scriptsize$\pm0.9$ & 73.3\scriptsize$\pm0.4$ & 
		9.9\scriptsize$\pm0.3$ & 
		51.7 &
		
		95.1\scriptsize{$\pm0.6$} & 98.7\scriptsize{$\pm0.2$} & 
		98.7\scriptsize{$\pm0.2$} & 98.7\scriptsize{$\pm0.2$} & 
		98.8\scriptsize{$\pm0.1$} & 95.6\scriptsize{$\pm0.4$} & 97.6 \\
		
		ARM \cite{zhang2020arm}  & 
		74.5\scriptsize{$\pm3.8$} & 71.1\scriptsize$\pm1.8$ & 
		9.9\scriptsize$\pm0.3$   & 
		51.8 &
		
		95.1\scriptsize{$\pm1.1$} & 98.8\scriptsize{$\pm0.2$} & 
		98.8\scriptsize{$\pm0.1$} & 98.8\scriptsize{$\pm0.1$} & 
		98.8\scriptsize{$\pm0.1$} & 96\scriptsize{$\pm0.6$} & 97.7 \\ 
		
		CORAL \cite{sun2016coral} & 
		72.3\scriptsize$\pm0.7$ & 72.8\scriptsize$\pm0.4$ & 
		10.5\scriptsize$\pm0.3$ & 
		51.8 &
		
		95.6\scriptsize{$\pm0.3$} & 98.9\scriptsize{$\pm0.1$} & 
		98.9\scriptsize{$\pm0.1$} & 99.0\scriptsize{$\pm0.0$} & 
		98.9\scriptsize{$\pm0.1$} & 96.1\scriptsize{$\pm0.3$} & 97.9 \\

		Fish \cite{shi2021gradient}  & 
		71.7\scriptsize$\pm0.5$ & 73.2\scriptsize$\pm0.5$ & 
		10.4\scriptsize$\pm0.2$ & 
		51.8 &
		95.3\scriptsize{$\pm0.6$} & 98.9\scriptsize{$\pm0.1$} & 
		98.9\scriptsize{$\pm0.2$} & 98.9\scriptsize{$\pm0.1$} & 
		98.9\scriptsize{$\pm0.1$} & 95.6\scriptsize{$\pm0.6$} & 97.7 \\ 
		
		GroupDRO \cite{Sagawa2020GroupDRO} & 
		72.6\scriptsize$\pm0.6$ & 73.5\scriptsize$\pm0.4$ & 
		9.9\scriptsize$\pm0.2$ & 
		52.0 &
		95.9\scriptsize{$\pm0.6$} & 98.7\scriptsize{$\pm0.2$} & 
		98.6\scriptsize{$\pm0.1$} & 98.7\scriptsize{$\pm0.1$} & 
		98.7\scriptsize{$\pm0.1$} & 96.0\scriptsize{$\pm0.2$} & 97.8\\  	
		
		VREx \cite{krueger2020vrex} & 
		72.9\scriptsize$\pm0.3$ & 72.9\scriptsize$\pm0.4$ & 
		10.3\scriptsize$\pm0.6$ & 
		52.0 &
		95.7\scriptsize{$\pm0.6$} & 98.9\scriptsize{$\pm0.2$} & 
		98.7\scriptsize{$\pm0.1$} & 98.9\scriptsize{$\pm0.2$} & 
		98.9\scriptsize{$\pm0.1$} & 95.8\scriptsize{$\pm0.4$} & 97.8\\

		\midrule
		
		SAM \cite{foret2020sharpness} & 
		71.1\scriptsize$\pm0.5$ & 73.3\scriptsize$\pm0.4$ & 
		10.1\scriptsize$\pm0.3$ & 
		51.5 &
		95.7\scriptsize{$\pm0.2$} & 99.0\scriptsize{$\pm0.1$} & 
		98.9\scriptsize{$\pm0.1$} & 98.9\scriptsize{$\pm0.1$} & 
		98.9\scriptsize{$\pm0.1$} & 96.2\scriptsize{$\pm0.4$} & 97.9\\
		
		GSAM \cite{zhuang2022surrogate} & 
		71.8\scriptsize$\pm0.3$ & 73.2\scriptsize$\pm0.2$ & 
		9.9\scriptsize$\pm0.2$ & 
		51.6 &
		94.9\scriptsize{$\pm0.1$} & 98.9\scriptsize{$\pm0.1$} & 
		98.9\scriptsize{$\pm0.2$} & 99.0\scriptsize{$\pm0.2$} & 
		98.8\scriptsize{$\pm0.1$} & 96.0\scriptsize{$\pm0.1$} & 97.7\\
		
		RSC \cite{huang2020rsc} & 
		72.5\scriptsize$\pm0.3$ & 72.4\scriptsize$\pm0.6$ & 
		10.2\scriptsize$\pm0.5$ & 
		51.7 &
		94.2\scriptsize{$\pm1.1$} & 98.6\scriptsize{$\pm0.1$} & 
		98.7\scriptsize{$\pm0.2$} & 98.6\scriptsize{$\pm0.2$} & 
		98.7\scriptsize{$\pm0.2$} & 95.7\scriptsize{$\pm0.7$} & 97.4\\

		SAGM \cite{wang2023sharpness} & 
		71.5\scriptsize$\pm0.8$ & 73.6\scriptsize$\pm0.5$ & 
		10.6\scriptsize$\pm0.6$ & 
		51.9 &
		95.4\scriptsize{$\pm0.4$} & 98.9\scriptsize{$\pm0.1$} & 
		98.9\scriptsize{$\pm0.1$} & 98.9\scriptsize{$\pm0.1$} & 
		98.9\scriptsize{$\pm0.1$} & 95.9\scriptsize{$\pm0.5$} & 97.8\\
		
		\midrule
		\textbf{GGA} (ours) & 
		71.2\scriptsize$\pm0.7$ & 73.1\scriptsize$\pm0.6$ & 
		11.5\scriptsize$\pm0.4$ & 
		51.9 &
		95.1\scriptsize{$\pm0.8$} & 99.0\scriptsize{$\pm0.1$} & 
		99.0\scriptsize{$\pm0.3$} & 98.8\scriptsize{$\pm0.1$} & 
		98.8\scriptsize{$\pm0.2$} & 96.1\scriptsize{$\pm0.4$} & 97.8 \\ 
		
		\textbf{GGA-L} (ours) & 
		71.6\scriptsize$\pm0.3$ & 
		73.1\scriptsize$\pm0.2$ & 
		11.0\scriptsize$\pm0.9$ & 
		51.9 &
		95.4\scriptsize{$\pm0.4$} & 98.6\scriptsize{$\pm0.2$} & 
		98.7\scriptsize{$\pm0.3$} & 98.8\scriptsize{$\pm0.1$} & 
		98.7\scriptsize{$\pm0.2$} & 95.5\scriptsize{$\pm0.4$} & 97.6 \\ 
		\bottomrule
	\end{tabular}
\end{table*}

\clearpage
\clearpage

\bibliographystyle{ieeenat_fullname}
\bibliography{supp}

\begin{thebibliography}{99}
\providecommand{\natexlab}[1]{#1}
\providecommand{\url}[1]{\texttt{#1}}
\expandafter\ifx\csname urlstyle\endcsname\relax
  \providecommand{\doi}[1]{doi: #1}\else
  \providecommand{\doi}{doi: \begingroup \urlstyle{rm}\Url}\fi

\bibitem[Arjovsky et~al.(2019)Arjovsky, Bottou, Gulrajani, and
  Lopez-Paz]{arjovsky2019irm}
Martin Arjovsky, L{\'e}on Bottou, Ishaan Gulrajani, and David Lopez-Paz.
\newblock Invariant risk minimization.
\newblock \emph{arXiv preprint arXiv:1907.02893}, 2019.

\bibitem[Auer et~al.(1995)Auer, Herbster, and Warmuth]{auer1995exponentially}
Peter Auer, Mark Herbster, and Manfred~KK Warmuth.
\newblock Exponentially many local minima for single neurons.
\newblock \emph{Advances in neural information processing systems}, 8, 1995.

\bibitem[Balaji et~al.(2018)Balaji, Sankaranarayanan, and
  Chellappa]{balaji2018metareg}
Yogesh Balaji, Swami Sankaranarayanan, and Rama Chellappa.
\newblock Metareg: Towards domain generalization using meta-regularization.
\newblock \emph{Advances in neural information processing systems}, 31, 2018.

\bibitem[Ballas and Diou(2024{\natexlab{a}})]{10233054}
Aristotelis Ballas and Christos Diou.
\newblock Towards domain generalization for ecg and eeg classification:
  Algorithms and benchmarks.
\newblock \emph{IEEE Transactions on Emerging Topics in Computational
  Intelligence}, 8\penalty0 (1):\penalty0 44--54, 2024{\natexlab{a}}.

\bibitem[Ballas and Diou(2024{\natexlab{b}})]{10472869}
Aristotelis Ballas and Christos Diou.
\newblock Multi-scale and multi-layer contrastive learning for domain
  generalization.
\newblock \emph{IEEE Transactions on Artificial Intelligence}, pages 1--14,
  2024{\natexlab{b}}.

\bibitem[Ballas and Diou(2024{\natexlab{c}})]{ballas2024multi}
Aristotelis Ballas and Christos Diou.
\newblock Multi-scale and multi-layer contrastive learning for domain
  generalization.
\newblock \emph{IEEE Transactions on Artificial Intelligence},
  2024{\natexlab{c}}.

\bibitem[Beery et~al.(2018)Beery, Van~Horn, and Perona]{beery2018recognition}
Sara Beery, Grant Van~Horn, and Pietro Perona.
\newblock Recognition in terra incognita.
\newblock In \emph{Proceedings of the European conference on computer vision
  (ECCV)}, pages 456--473, 2018.

\bibitem[Ben-David et~al.(2006)Ben-David, Blitzer, Crammer, and
  Pereira]{ben2006analysis}
Shai Ben-David, John Blitzer, Koby Crammer, and Fernando Pereira.
\newblock Analysis of representations for domain adaptation.
\newblock \emph{Advances in neural information processing systems}, 19, 2006.

\bibitem[Bengio et~al.(2013)Bengio, Courville, and Vincent]{6472238}
Yoshua Bengio, Aaron Courville, and Pascal Vincent.
\newblock Representation learning: A review and new perspectives.
\newblock \emph{IEEE Transactions on Pattern Analysis and Machine
  Intelligence}, 35\penalty0 (8):\penalty0 1798--1828, 2013.

\bibitem[Blanchard et~al.(2021)Blanchard, Deshmukh, Dogan, Lee, and
  Scott]{blanchard2021mtl_marginal_transfer_learning}
Gilles Blanchard, Aniket~Anand Deshmukh, Urun Dogan, Gyemin Lee, and Clayton
  Scott.
\newblock Domain generalization by marginal transfer learning.
\newblock \emph{Journal of Machine Learning Research}, 22\penalty0
  (2):\penalty0 1--55, 2021.

\bibitem[Borlino et~al.(2021)Borlino, D'Innocente, and
  Tommasi]{borlino2021rethinking}
Francesco~Cappio Borlino, Antonio D'Innocente, and Tatiana Tommasi.
\newblock Rethinking domain generalization baselines.
\newblock In \emph{2020 25th International Conference on Pattern Recognition
  (ICPR)}, pages 9227--9233. IEEE, 2021.

\bibitem[Bottou(1998)]{bottou1998online}
L{\'e}on Bottou.
\newblock Online algorithms and stochastic approximations.
\newblock \emph{Online learning in neural networks}, 1998.

\bibitem[Cai(2021)]{cai2021sa}
Zhicheng Cai.
\newblock Sa-gd: Improved gradient descent learning strategy with simulated
  annealing.
\newblock \emph{arXiv preprint arXiv:2107.07558}, 2021.

\bibitem[Carlucci et~al.(2019)Carlucci, D'Innocente, Bucci, Caputo, and
  Tommasi]{carlucci2019domain}
Fabio~M Carlucci, Antonio D'Innocente, Silvia Bucci, Barbara Caputo, and
  Tatiana Tommasi.
\newblock Domain generalization by solving jigsaw puzzles.
\newblock In \emph{Proceedings of the IEEE/CVF conference on computer vision
  and pattern recognition}, pages 2229--2238, 2019.

\bibitem[Cha et~al.(2021)Cha, Chun, Lee, Cho, Park, Lee, and Park]{cha2021swad}
Junbum Cha, Sanghyuk Chun, Kyungjae Lee, Han-Cheol Cho, Seunghyun Park, Yunsung
  Lee, and Sungrae Park.
\newblock Swad: Domain generalization by seeking flat minima.
\newblock \emph{Advances in Neural Information Processing Systems},
  34:\penalty0 22405--22418, 2021.

\bibitem[Chattopadhyay et~al.(2020)Chattopadhyay, Balaji, and
  Hoffman]{chattopadhyay2020learning}
Prithvijit Chattopadhyay, Yogesh Balaji, and Judy Hoffman.
\newblock Learning to balance specificity and invariance for in and out of
  domain generalization.
\newblock In \emph{Computer Vision--ECCV 2020: 16th European Conference,
  Glasgow, UK, August 23--28, 2020, Proceedings, Part IX 16}, pages 301--318.
  Springer, 2020.

\bibitem[Chen et~al.(2016)Chen, Duan, Houthooft, Schulman, Sutskever, and
  Abbeel]{chen2016infogan}
Xi Chen, Yan Duan, Rein Houthooft, John Schulman, Ilya Sutskever, and Pieter
  Abbeel.
\newblock Infogan: Interpretable representation learning by information
  maximizing generative adversarial nets.
\newblock \emph{Advances in neural information processing systems}, 29, 2016.

\bibitem[Chen et~al.(2018)Chen, Badrinarayanan, Lee, and
  Rabinovich]{chen2018gradnorm}
Zhao Chen, Vijay Badrinarayanan, Chen-Yu Lee, and Andrew Rabinovich.
\newblock Gradnorm: Gradient normalization for adaptive loss balancing in deep
  multitask networks.
\newblock In \emph{International conference on machine learning}, pages
  794--803. PMLR, 2018.

\bibitem[Choi et~al.(2023)Choi, Das, Choi, Yang, Park, and
  Yun]{choi2023progressive}
Seokeon Choi, Debasmit Das, Sungha Choi, Seunghan Yang, Hyunsin Park, and
  Sungrack Yun.
\newblock Progressive random convolutions for single domain generalization.
\newblock In \emph{Proceedings of the IEEE/CVF Conference on Computer Vision
  and Pattern Recognition}, pages 10312--10322, 2023.

\bibitem[Choromanska et~al.(2015)Choromanska, Henaff, Mathieu, Arous, and
  LeCun]{choromanska2015loss}
Anna Choromanska, Mikael Henaff, Michael Mathieu, G{\'e}rard~Ben Arous, and
  Yann LeCun.
\newblock The loss surfaces of multilayer networks.
\newblock In \emph{Artificial intelligence and statistics}, pages 192--204.
  PMLR, 2015.

\bibitem[Correia et~al.(2023)Correia, Worrall, and
  Bondesan]{pmlr-v206-correia23a}
Alvaro~H.C. Correia, Daniel~E. Worrall, and Roberto Bondesan.
\newblock Neural simulated annealing.
\newblock In \emph{Proceedings of The 26th International Conference on
  Artificial Intelligence and Statistics}, pages 4946--4962. PMLR, 2023.

\bibitem[Ding and Fu(2017)]{ding2017deep}
Zhengming Ding and Yun Fu.
\newblock Deep domain generalization with structured low-rank constraint.
\newblock \emph{IEEE Transactions on Image Processing}, 27\penalty0
  (1):\penalty0 304--313, 2017.

\bibitem[Dissanayake et~al.(2021)Dissanayake, Fernando, Denman, Ghaemmaghami,
  Sridharan, and Fookes]{9298838}
Theekshana Dissanayake, Tharindu Fernando, Simon Denman, Houman Ghaemmaghami,
  Sridha Sridharan, and Clinton Fookes.
\newblock Domain generalization in biosignal classification.
\newblock \emph{IEEE Transactions on Biomedical Engineering}, 68\penalty0
  (6):\penalty0 1978--1989, 2021.

\bibitem[Dou et~al.(2019)Dou, Coelho~de Castro, Kamnitsas, and
  Glocker]{dou2019domain}
Qi Dou, Daniel Coelho~de Castro, Konstantinos Kamnitsas, and Ben Glocker.
\newblock Domain generalization via model-agnostic learning of semantic
  features.
\newblock \emph{Advances in neural information processing systems}, 32, 2019.

\bibitem[Du et~al.(2021)Du, Wang, Feng, Pan, Qin, Xu, and Wang]{du2021adarnn}
Yuntao Du, Jindong Wang, Wenjie Feng, Sinno Pan, Tao Qin, Renjun Xu, and
  Chongjun Wang.
\newblock Adarnn: Adaptive learning and forecasting of time series.
\newblock In \emph{Proceedings of the 30th ACM international conference on
  information \& knowledge management}, pages 402--411, 2021.

\bibitem[Fang et~al.(2013)Fang, Xu, and Rockmore]{fang2013unbiased}
Chen Fang, Ye Xu, and Daniel~N Rockmore.
\newblock Unbiased metric learning: On the utilization of multiple datasets and
  web images for softening bias.
\newblock In \emph{Proceedings of the IEEE International Conference on Computer
  Vision}, pages 1657--1664, 2013.

\bibitem[Finn et~al.(2017)Finn, Abbeel, and Levine]{finn2017model}
Chelsea Finn, Pieter Abbeel, and Sergey Levine.
\newblock Model-agnostic meta-learning for fast adaptation of deep networks.
\newblock In \emph{International conference on machine learning}, pages
  1126--1135. PMLR, 2017.

\bibitem[Foret et~al.(2021)Foret, Kleiner, Mobahi, and
  Neyshabur]{foret2020sharpness}
Pierre Foret, Ariel Kleiner, Hossein Mobahi, and Behnam Neyshabur.
\newblock Sharpness-aware minimization for efficiently improving
  generalization.
\newblock In \emph{International Conference on Learning Representations}, 2021.

\bibitem[Ganin et~al.(2016{\natexlab{a}})Ganin, Ustinova, Ajakan, Germain,
  Larochelle, Laviolette, March, and Lempitsky]{ganin2016domain}
Yaroslav Ganin, Evgeniya Ustinova, Hana Ajakan, Pascal Germain, Hugo
  Larochelle, Fran{\c{c}}ois Laviolette, Mario March, and Victor Lempitsky.
\newblock Domain-adversarial training of neural networks.
\newblock \emph{Journal of machine learning research}, 17\penalty0
  (59):\penalty0 1--35, 2016{\natexlab{a}}.

\bibitem[Ganin et~al.(2016{\natexlab{b}})Ganin, Ustinova, Ajakan, Germain,
  Larochelle, Laviolette, Marchand, and Lempitsky]{ganin2016dann}
Yaroslav Ganin, Evgeniya Ustinova, Hana Ajakan, Pascal Germain, Hugo
  Larochelle, Fran{\c{c}}ois Laviolette, Mario Marchand, and Victor Lempitsky.
\newblock Domain-adversarial training of neural networks.
\newblock \emph{Journal of machine learning research}, 17\penalty0
  (1):\penalty0 2096--2030, 2016{\natexlab{b}}.

\bibitem[Gulrajani and Lopez-Paz(2021)]{gulrajani2020domainbed}
Ishaan Gulrajani and David Lopez-Paz.
\newblock In search of lost domain generalization.
\newblock In \emph{International Conference on Learning Representations}, 2021.

\bibitem[He et~al.(2016)He, Zhang, Ren, and Sun]{he2016deep}
Kaiming He, Xiangyu Zhang, Shaoqing Ren, and Jian Sun.
\newblock Deep residual learning for image recognition.
\newblock In \emph{Proceedings of the IEEE conference on computer vision and
  pattern recognition}, pages 770--778, 2016.

\bibitem[Huang and Belongie(2017)]{huang2017arbitrary}
Xun Huang and Serge Belongie.
\newblock Arbitrary style transfer in real-time with adaptive instance
  normalization.
\newblock In \emph{Proceedings of the IEEE international conference on computer
  vision}, pages 1501--1510, 2017.

\bibitem[Huang et~al.(2020)Huang, Wang, Xing, and Huang]{huang2020rsc}
Zeyi Huang, Haohan Wang, Eric~P Xing, and Dong Huang.
\newblock Self-challenging improves cross-domain generalization.
\newblock \emph{European Conference on Computer Vision}, 2020.

\bibitem[Hupkes et~al.(2023)Hupkes, Giulianelli, Dankers, Artetxe, Elazar,
  Pimentel, Christodoulopoulos, Lasri, Saphra, Sinclair,
  et~al.]{hupkes2023taxonomy}
Dieuwke Hupkes, Mario Giulianelli, Verna Dankers, Mikel Artetxe, Yanai Elazar,
  Tiago Pimentel, Christos Christodoulopoulos, Karim Lasri, Naomi Saphra,
  Arabella Sinclair, et~al.
\newblock A taxonomy and review of generalization research in nlp.
\newblock \emph{Nature Machine Intelligence}, 5\penalty0 (10):\penalty0
  1161--1174, 2023.

\bibitem[Ilse et~al.(2020)Ilse, Tomczak, Louizos, and Welling]{ilse2020diva}
Maximilian Ilse, Jakub~M Tomczak, Christos Louizos, and Max Welling.
\newblock Diva: Domain invariant variational autoencoders.
\newblock In \emph{Medical Imaging with Deep Learning}, pages 322--348. PMLR,
  2020.

\bibitem[Izmailov et~al.(2018)Izmailov, Podoprikhin, Garipov, Vetrov, and
  Wilson]{izmailov2018averaging}
Pavel Izmailov, Dmitrii Podoprikhin, Timur Garipov, Dmitry Vetrov, and
  Andrew~Gordon Wilson.
\newblock Averaging weights leads to wider optima and better generalization.
\newblock \emph{arXiv preprint arXiv:1803.05407}, 2018.

\bibitem[Kendall et~al.(2018)Kendall, Gal, and Cipolla]{kendall2018multi}
Alex Kendall, Yarin Gal, and Roberto Cipolla.
\newblock Multi-task learning using uncertainty to weigh losses for scene
  geometry and semantics.
\newblock In \emph{Proceedings of the IEEE conference on computer vision and
  pattern recognition}, pages 7482--7491, 2018.

\bibitem[Kim et~al.(2021)Kim, Yoo, Park, Kim, and Lee]{kim2021selfreg}
Daehee Kim, Youngjun Yoo, Seunghyun Park, Jinkyu Kim, and Jaekoo Lee.
\newblock Selfreg: Self-supervised contrastive regularization for domain
  generalization.
\newblock In \emph{International Conference on Computer Vision}, 2021.

\bibitem[Kingma(2014)]{kingma2014adam}
Diederik~P Kingma.
\newblock Adam: A method for stochastic optimization.
\newblock \emph{arXiv preprint arXiv:1412.6980}, 2014.

\bibitem[Kirkpatrick et~al.(1983)Kirkpatrick, Gelatt~Jr, and
  Vecchi]{kirkpatrick1983optimization}
Scott Kirkpatrick, C~Daniel Gelatt~Jr, and Mario~P Vecchi.
\newblock Optimization by simulated annealing.
\newblock \emph{science}, 220\penalty0 (4598):\penalty0 671--680, 1983.

\bibitem[Krueger et~al.(2021{\natexlab{a}})Krueger, Caballero, Jacobsen, Zhang,
  Binas, Zhang, Le~Priol, and Courville]{krueger2020vrex}
David Krueger, Ethan Caballero, Joern-Henrik Jacobsen, Amy Zhang, Jonathan
  Binas, Dinghuai Zhang, Remi Le~Priol, and Aaron Courville.
\newblock Out-of-distribution generalization via risk extrapolation (rex).
\newblock In \emph{International conference on machine learning}, pages
  5815--5826. PMLR, 2021{\natexlab{a}}.

\bibitem[Krueger et~al.(2021{\natexlab{b}})Krueger, Caballero, Jacobsen, Zhang,
  Binas, Zhang, Le~Priol, and Courville]{krueger2021out}
David Krueger, Ethan Caballero, Joern-Henrik Jacobsen, Amy Zhang, Jonathan
  Binas, Dinghuai Zhang, Remi Le~Priol, and Aaron Courville.
\newblock Out-of-distribution generalization via risk extrapolation (rex).
\newblock In \emph{International conference on machine learning}, pages
  5815--5826. PMLR, 2021{\natexlab{b}}.

\bibitem[Laskin et~al.(2020)Laskin, Srinivas, and Abbeel]{laskin2020curl}
Michael Laskin, Aravind Srinivas, and Pieter Abbeel.
\newblock Curl: Contrastive unsupervised representations for reinforcement
  learning.
\newblock In \emph{International conference on machine learning}, pages
  5639--5650. PMLR, 2020.

\bibitem[Le and Woo(2024)]{le2024gradient}
Binh~M Le and Simon~S Woo.
\newblock Gradient alignment for cross-domain face anti-spoofing.
\newblock In \emph{Proceedings of the IEEE/CVF Conference on Computer Vision
  and Pattern Recognition}, pages 188--199, 2024.

\bibitem[Lee et~al.(2019)Lee, Lee, Shin, and Lee]{lee2019network}
Kimin Lee, Kibok Lee, Jinwoo Shin, and Honglak Lee.
\newblock Network randomization: A simple technique for generalization in deep
  reinforcement learning.
\newblock \emph{arXiv preprint arXiv:1910.05396}, 2019.

\bibitem[Li et~al.(2017)Li, Yang, Song, and Hospedales]{li2017deeper}
Da Li, Yongxin Yang, Yi-Zhe Song, and Timothy~M Hospedales.
\newblock Deeper, broader and artier domain generalization.
\newblock In \emph{Proceedings of the IEEE international conference on computer
  vision}, pages 5542--5550, 2017.

\bibitem[Li et~al.(2018{\natexlab{a}})Li, Yang, Song, and
  Hospedales]{li2018learning}
Da Li, Yongxin Yang, Yi-Zhe Song, and Timothy Hospedales.
\newblock Learning to generalize: Meta-learning for domain generalization.
\newblock In \emph{Proceedings of the AAAI conference on artificial
  intelligence}, 2018{\natexlab{a}}.

\bibitem[Li et~al.(2018{\natexlab{b}})Li, Pan, Wang, and Kot]{li2018mmd}
Haoliang Li, Sinno~Jialin Pan, Shiqi Wang, and Alex~C Kot.
\newblock Domain generalization with adversarial feature learning.
\newblock In \emph{Computer Vision and Pattern Recognition},
  2018{\natexlab{b}}.

\bibitem[Li et~al.(2018{\natexlab{c}})Li, Tian, Gong, Liu, Liu, Zhang, and
  Tao]{li2018deep}
Ya Li, Xinmei Tian, Mingming Gong, Yajing Liu, Tongliang Liu, Kun Zhang, and
  Dacheng Tao.
\newblock Deep domain generalization via conditional invariant adversarial
  networks.
\newblock In \emph{Proceedings of the European conference on computer vision
  (ECCV)}, pages 624--639, 2018{\natexlab{c}}.

\bibitem[Lv et~al.(2022)Lv, Liang, Li, Zang, Liu, Wang, and
  Liu]{lv2022causality}
Fangrui Lv, Jian Liang, Shuang Li, Bin Zang, Chi~Harold Liu, Ziteng Wang, and
  Di Liu.
\newblock Causality inspired representation learning for domain generalization.
\newblock In \emph{Proceedings of the IEEE/CVF conference on computer vision
  and pattern recognition}, pages 8046--8056, 2022.

\bibitem[Mahajan et~al.(2021)Mahajan, Tople, and Sharma]{mahajan2021domain}
Divyat Mahajan, Shruti Tople, and Amit Sharma.
\newblock Domain generalization using causal matching.
\newblock In \emph{International conference on machine learning}, pages
  7313--7324. PMLR, 2021.

\bibitem[Mansilla et~al.(2021)Mansilla, Echeveste, Milone, and
  Ferrante]{mansilla2021domain}
Lucas Mansilla, Rodrigo Echeveste, Diego~H Milone, and Enzo Ferrante.
\newblock Domain generalization via gradient surgery.
\newblock In \emph{Proceedings of the IEEE/CVF international conference on
  computer vision}, pages 6630--6638, 2021.

\bibitem[Matsuura and Harada(2020)]{matsuura2020domain}
Toshihiko Matsuura and Tatsuya Harada.
\newblock Domain generalization using a mixture of multiple latent domains.
\newblock In \emph{Proceedings of the AAAI conference on artificial
  intelligence}, pages 11749--11756, 2020.

\bibitem[Muandet et~al.(2013)Muandet, Balduzzi, and
  Sch{\"o}lkopf]{muandet2013domain}
Krikamol Muandet, David Balduzzi, and Bernhard Sch{\"o}lkopf.
\newblock Domain generalization via invariant feature representation.
\newblock In \emph{International conference on machine learning}, pages 10--18.
  PMLR, 2013.

\bibitem[Nam et~al.(2021)Nam, Lee, Park, Yoon, and Yoo]{nam2019sagnet}
Hyeonseob Nam, HyunJae Lee, Jongchan Park, Wonjun Yoon, and Donggeun Yoo.
\newblock Reducing domain gap by reducing style bias.
\newblock In \emph{Computer Vision and Pattern Recognition}, 2021.

\bibitem[Nguyen and Hein(2018)]{nguyen2018optimization}
Quynh Nguyen and Matthias Hein.
\newblock Optimization landscape and expressivity of deep cnns.
\newblock In \emph{International conference on machine learning}, pages
  3730--3739. PMLR, 2018.

\bibitem[Ot{\'a}lora et~al.(2019)Ot{\'a}lora, Atzori, Andrearczyk, Khan, and
  M{\"u}ller]{otalora2019staining}
Sebastian Ot{\'a}lora, Manfredo Atzori, Vincent Andrearczyk, Amjad Khan, and
  Henning M{\"u}ller.
\newblock Staining invariant features for improving generalization of deep
  convolutional neural networks in computational pathology.
\newblock \emph{Frontiers in bioengineering and biotechnology}, 7:\penalty0
  198, 2019.

\bibitem[Peng et~al.(2019{\natexlab{a}})Peng, Bai, Xia, Huang, Saenko, and
  Wang]{peng2019moment}
Xingchao Peng, Qinxun Bai, Xide Xia, Zijun Huang, Kate Saenko, and Bo Wang.
\newblock Moment matching for multi-source domain adaptation.
\newblock In \emph{Proceedings of the IEEE/CVF international conference on
  computer vision}, pages 1406--1415, 2019{\natexlab{a}}.

\bibitem[Peng et~al.(2019{\natexlab{b}})Peng, Huang, Sun, and
  Saenko]{peng2019domain}
Xingchao Peng, Zijun Huang, Ximeng Sun, and Kate Saenko.
\newblock Domain agnostic learning with disentangled representations.
\newblock In \emph{International conference on machine learning}, pages
  5102--5112. PMLR, 2019{\natexlab{b}}.

\bibitem[Piratla et~al.(2020)Piratla, Netrapalli, and
  Sarawagi]{piratla2020efficient}
Vihari Piratla, Praneeth Netrapalli, and Sunita Sarawagi.
\newblock Efficient domain generalization via common-specific low-rank
  decomposition.
\newblock In \emph{International conference on machine learning}, pages
  7728--7738. PMLR, 2020.

\bibitem[Qiao et~al.(2020)Qiao, Zhao, and Peng]{qiao2020learning}
Fengchun Qiao, Long Zhao, and Xi Peng.
\newblock Learning to learn single domain generalization.
\newblock In \emph{Proceedings of the IEEE/CVF conference on computer vision
  and pattern recognition}, pages 12556--12565, 2020.

\bibitem[Rere et~al.(2015)Rere, Fanany, and Arymurthy]{RERE2015137}
L.M.~Rasdi Rere, Mohamad~Ivan Fanany, and Aniati~Murni Arymurthy.
\newblock Simulated annealing algorithm for deep learning.
\newblock \emph{Procedia Computer Science}, 72:\penalty0 137--144, 2015.
\newblock The Third Information Systems International Conference 2015.

\bibitem[Russakovsky et~al.(2015)Russakovsky, Deng, Su, Krause, Satheesh, Ma,
  Huang, Karpathy, Khosla, Bernstein, et~al.]{russakovsky2015imagenet}
Olga Russakovsky, Jia Deng, Hao Su, Jonathan Krause, Sanjeev Satheesh, Sean Ma,
  Zhiheng Huang, Andrej Karpathy, Aditya Khosla, Michael Bernstein, et~al.
\newblock Imagenet large scale visual recognition challenge.
\newblock \emph{International journal of computer vision}, 115:\penalty0
  211--252, 2015.

\bibitem[Safran and Shamir(2016)]{safran2016quality}
Itay Safran and Ohad Shamir.
\newblock On the quality of the initial basin in overspecified neural networks.
\newblock In \emph{International Conference on Machine Learning}, pages
  774--782. PMLR, 2016.

\bibitem[Sagawa* et~al.(2020)Sagawa*, Koh*, Hashimoto, and
  Liang]{Sagawa2020GroupDRO}
Shiori Sagawa*, Pang~Wei Koh*, Tatsunori~B. Hashimoto, and Percy Liang.
\newblock Distributionally robust neural networks.
\newblock In \emph{International Conference on Learning Representations}, 2020.

\bibitem[Sarfi et~al.(2023)Sarfi, Karimpour, Chaudhary, Khalid, Ravanelli,
  Mudur, and Belilovsky]{sarfi2023simulated}
Amir~M Sarfi, Zahra Karimpour, Muawiz Chaudhary, Nasir~M Khalid, Mirco
  Ravanelli, Sudhir Mudur, and Eugene Belilovsky.
\newblock Simulated annealing in early layers leads to better generalization.
\newblock In \emph{Proceedings of the IEEE/CVF Conference on Computer Vision
  and Pattern Recognition}, pages 20205--20214, 2023.

\bibitem[Sener and Koltun(2018)]{sener2018multi}
Ozan Sener and Vladlen Koltun.
\newblock Multi-task learning as multi-objective optimization.
\newblock \emph{Advances in neural information processing systems}, 31, 2018.

\bibitem[Shahtalebi et~al.(2021)Shahtalebi, Gagnon-Audet, Laleh, Faramarzi,
  Ahuja, and Rish]{shahtalebi2021sand}
Soroosh Shahtalebi, Jean-Christophe Gagnon-Audet, Touraj Laleh, Mojtaba
  Faramarzi, Kartik Ahuja, and Irina Rish.
\newblock Sand-mask: An enhanced gradient masking strategy for the discovery of
  invariances in domain generalization.
\newblock \emph{arXiv preprint arXiv:2106.02266}, 2021.

\bibitem[Shankar et~al.(2018)Shankar, Piratla, Chakrabarti, Chaudhuri, Jyothi,
  and Sarawagi]{shankar2018generalizing}
Shiv Shankar, Vihari Piratla, Soumen Chakrabarti, Siddhartha Chaudhuri, Preethi
  Jyothi, and Sunita Sarawagi.
\newblock Generalizing across domains via cross-gradient training.
\newblock \emph{arXiv preprint arXiv:1804.10745}, 2018.

\bibitem[Shi et~al.(2022)Shi, Seely, Torr, N, Hannun, Usunier, and
  Synnaeve]{shi2021gradient}
Yuge Shi, Jeffrey Seely, Philip Torr, Siddharth N, Awni Hannun, Nicolas
  Usunier, and Gabriel Synnaeve.
\newblock Gradient matching for domain generalization.
\newblock In \emph{International Conference on Learning Representations}, 2022.

\bibitem[Sun and Saenko(2016)]{sun2016coral}
Baochen Sun and Kate Saenko.
\newblock Deep coral: Correlation alignment for deep domain adaptation.
\newblock In \emph{European Conference on Computer Vision}, 2016.

\bibitem[Tobin et~al.(2017)Tobin, Fong, Ray, Schneider, Zaremba, and
  Abbeel]{tobin2017domain}
Josh Tobin, Rachel Fong, Alex Ray, Jonas Schneider, Wojciech Zaremba, and
  Pieter Abbeel.
\newblock Domain randomization for transferring deep neural networks from
  simulation to the real world.
\newblock In \emph{2017 IEEE/RSJ international conference on intelligent robots
  and systems (IROS)}, pages 23--30. IEEE, 2017.

\bibitem[Vapnik(1998)]{vapnik1998statistical}
V Vapnik.
\newblock Statistical learning theory.
\newblock \emph{NY: Wiley}, 1998.

\bibitem[Venkateswara et~al.(2017)Venkateswara, Eusebio, Chakraborty, and
  Panchanathan]{venkateswara2017deep}
Hemanth Venkateswara, Jose Eusebio, Shayok Chakraborty, and Sethuraman
  Panchanathan.
\newblock Deep hashing network for unsupervised domain adaptation.
\newblock In \emph{Proceedings of the IEEE conference on computer vision and
  pattern recognition}, pages 5018--5027, 2017.

\bibitem[Volpi and Murino(2019)]{volpi2019addressing}
Riccardo Volpi and Vittorio Murino.
\newblock Addressing model vulnerability to distributional shifts over image
  transformation sets.
\newblock In \emph{Proceedings of the IEEE/CVF International Conference on
  Computer Vision}, pages 7980--7989, 2019.

\bibitem[Volpi et~al.(2018)Volpi, Namkoong, Sener, Duchi, Murino, and
  Savarese]{volpi2018generalizing}
Riccardo Volpi, Hongseok Namkoong, Ozan Sener, John~C Duchi, Vittorio Murino,
  and Silvio Savarese.
\newblock Generalizing to unseen domains via adversarial data augmentation.
\newblock \emph{Advances in neural information processing systems}, 31, 2018.

\bibitem[Wang et~al.(2020{\natexlab{a}})Wang, Han, Shan, and
  Chen]{wang2020cross}
Guoqing Wang, Hu Han, Shiguang Shan, and Xilin Chen.
\newblock Cross-domain face presentation attack detection via multi-domain
  disentangled representation learning.
\newblock In \emph{Proceedings of the IEEE/CVF conference on computer vision
  and pattern recognition}, pages 6678--6687, 2020{\natexlab{a}}.

\bibitem[Wang et~al.(2022)Wang, Lan, Liu, Ouyang, Qin, Lu, Chen, Zeng, and
  Philip]{wang2022generalizing}
Jindong Wang, Cuiling Lan, Chang Liu, Yidong Ouyang, Tao Qin, Wang Lu, Yiqiang
  Chen, Wenjun Zeng, and S~Yu Philip.
\newblock Generalizing to unseen domains: A survey on domain generalization.
\newblock \emph{IEEE transactions on knowledge and data engineering},
  35\penalty0 (8):\penalty0 8052--8072, 2022.

\bibitem[Wang et~al.(2024)Wang, Ren, Shen, Huang, and Zhu]{wang2024multi}
Jun Wang, He Ren, Changqing Shen, Weiguo Huang, and Zhongkui Zhu.
\newblock Multi-scale style generative and adversarial contrastive networks for
  single domain generalization fault diagnosis.
\newblock \emph{Reliability Engineering \& System Safety}, 243:\penalty0
  109879, 2024.

\bibitem[Wang et~al.(2023)Wang, Zhang, Lei, and Zhang]{wang2023sharpness}
Pengfei Wang, Zhaoxiang Zhang, Zhen Lei, and Lei Zhang.
\newblock Sharpness-aware gradient matching for domain generalization.
\newblock In \emph{Proceedings of the IEEE/CVF Conference on Computer Vision
  and Pattern Recognition}, pages 3769--3778, 2023.

\bibitem[Wang et~al.(2020{\natexlab{b}})Wang, Yu, Li, Yang, Fu, and
  Heng]{wang2020dofe}
Shujun Wang, Lequan Yu, Kang Li, Xin Yang, Chi-Wing Fu, and Pheng-Ann Heng.
\newblock Dofe: Domain-oriented feature embedding for generalizable fundus
  image segmentation on unseen datasets.
\newblock \emph{IEEE Transactions on Medical Imaging}, 39\penalty0
  (12):\penalty0 4237--4248, 2020{\natexlab{b}}.

\bibitem[Wang et~al.(2021)Wang, Loog, and Van~Gemert]{wang2021respecting}
Ziqi Wang, Marco Loog, and Jan Van~Gemert.
\newblock Respecting domain relations: Hypothesis invariance for domain
  generalization.
\newblock In \emph{2020 25th International Conference on Pattern Recognition
  (ICPR)}, pages 9756--9763. IEEE, 2021.

\bibitem[Welling and Teh(2011)]{welling2011bayesian}
Max Welling and Yee~W Teh.
\newblock Bayesian learning via stochastic gradient langevin dynamics.
\newblock In \emph{Proceedings of the 28th international conference on machine
  learning (ICML-11)}, pages 681--688, 2011.

\bibitem[Xu et~al.(2020)Xu, Zhang, Ni, Li, Wang, Tian, and
  Zhang]{xu2020interdomain_mixup_aaai}
Minghao Xu, Jian Zhang, Bingbing Ni, Teng Li, Chengjie Wang, Qi Tian, and
  Wenjun Zhang.
\newblock Adversarial domain adaptation with domain mixup.
\newblock In \emph{AAAI Conference on Artificial Intelligence}, 2020.

\bibitem[Yosinski et~al.(2014)Yosinski, Clune, Bengio, and
  Lipson]{yosinski2014transferable}
Jason Yosinski, Jeff Clune, Yoshua Bengio, and Hod Lipson.
\newblock How transferable are features in deep neural networks?
\newblock \emph{Advances in neural information processing systems}, 27, 2014.

\bibitem[Yu et~al.(2020)Yu, Kumar, Gupta, Levine, Hausman, and
  Finn]{yu2020gradient}
Tianhe Yu, Saurabh Kumar, Abhishek Gupta, Sergey Levine, Karol Hausman, and
  Chelsea Finn.
\newblock Gradient surgery for multi-task learning.
\newblock \emph{Advances in Neural Information Processing Systems},
  33:\penalty0 5824--5836, 2020.

\bibitem[Yue et~al.(2019)Yue, Zhang, Zhao, Sangiovanni-Vincentelli, Keutzer,
  and Gong]{yue2019domain}
Xiangyu Yue, Yang Zhang, Sicheng Zhao, Alberto Sangiovanni-Vincentelli, Kurt
  Keutzer, and Boqing Gong.
\newblock Domain randomization and pyramid consistency: Simulation-to-real
  generalization without accessing target domain data.
\newblock In \emph{Proceedings of the IEEE/CVF international conference on
  computer vision}, pages 2100--2110, 2019.

\bibitem[Zhang et~al.(2022{\natexlab{a}})Zhang, Zhang, Liu, Weller,
  Sch\"olkopf, and Xing]{Zhang_2022_CVPR}
Hanlin Zhang, Yi-Fan Zhang, Weiyang Liu, Adrian Weller, Bernhard Sch\"olkopf,
  and Eric~P. Xing.
\newblock Towards principled disentanglement for domain generalization.
\newblock In \emph{Proceedings of the IEEE/CVF Conference on Computer Vision
  and Pattern Recognition (CVPR)}, pages 8024--8034, 2022{\natexlab{a}}.

\bibitem[Zhang et~al.(2022{\natexlab{b}})Zhang, Zhang, Liu, Weller,
  Sch{\"o}lkopf, and Xing]{zhang2022towards}
Hanlin Zhang, Yi-Fan Zhang, Weiyang Liu, Adrian Weller, Bernhard Sch{\"o}lkopf,
  and Eric~P Xing.
\newblock Towards principled disentanglement for domain generalization.
\newblock In \emph{Proceedings of the IEEE/CVF conference on computer vision
  and pattern recognition}, pages 8024--8034, 2022{\natexlab{b}}.

\bibitem[Zhang et~al.(2020{\natexlab{a}})Zhang, Wang, Yang, Sanford, Harmon,
  Turkbey, Wood, Roth, Myronenko, Xu, et~al.]{zhang2020generalizing}
Ling Zhang, Xiaosong Wang, Dong Yang, Thomas Sanford, Stephanie Harmon, Baris
  Turkbey, Bradford~J Wood, Holger Roth, Andriy Myronenko, Daguang Xu, et~al.
\newblock Generalizing deep learning for medical image segmentation to unseen
  domains via deep stacked transformation.
\newblock \emph{IEEE transactions on medical imaging}, 39\penalty0
  (7):\penalty0 2531--2540, 2020{\natexlab{a}}.

\bibitem[Zhang et~al.(2020{\natexlab{b}})Zhang, Marklund, Gupta, Levine, and
  Finn]{zhang2020arm}
Marvin Zhang, Henrik Marklund, Abhishek Gupta, Sergey Levine, and Chelsea Finn.
\newblock Adaptive risk minimization: A meta-learning approach for tackling
  group shift.
\newblock \emph{arXiv preprint arXiv:2007.02931}, 2020{\natexlab{b}}.

\bibitem[Zhang and Yang(2022)]{9392366}
Yu Zhang and Qiang Yang.
\newblock A survey on multi-task learning.
\newblock \emph{IEEE Transactions on Knowledge and Data Engineering},
  34\penalty0 (12):\penalty0 5586--5609, 2022.

\bibitem[Zheng et~al.(2021)Zheng, Yang, Yin, Li, Wang, and Xu]{9174912}
Huailiang Zheng, Yuantao Yang, Jiancheng Yin, Yuqing Li, Rixin Wang, and
  Minqiang Xu.
\newblock Deep domain generalization combining a priori diagnosis knowledge
  toward cross-domain fault diagnosis of rolling bearing.
\newblock \emph{IEEE Transactions on Instrumentation and Measurement},
  70:\penalty0 1--11, 2021.

\bibitem[Zhou et~al.(2021{\natexlab{a}})Zhou, Yang, Qiao, and
  Xiang]{zhou2021domain}
Kaiyang Zhou, Yongxin Yang, Yu Qiao, and Tao Xiang.
\newblock Domain adaptive ensemble learning.
\newblock \emph{IEEE Transactions on Image Processing}, 30:\penalty0
  8008--8018, 2021{\natexlab{a}}.

\bibitem[Zhou et~al.(2021{\natexlab{b}})Zhou, Yang, Qiao, and
  Xiang]{zhou2021mixstyle}
Kaiyang Zhou, Yongxin Yang, Yu Qiao, and Tao Xiang.
\newblock Domain generalization with mixstyle.
\newblock In \emph{International Conference on Learning Representations},
  2021{\natexlab{b}}.

\bibitem[Zhou et~al.(2023)Zhou, Liu, Qiao, Xiang, and Loy]{zhou_domain_2023}
Kaiyang Zhou, Ziwei Liu, Yu Qiao, Tao Xiang, and Chen~Change Loy.
\newblock Domain {Generalization}: {A} {Survey}.
\newblock \emph{IEEE Transactions on Pattern Analysis and Machine
  Intelligence}, 45\penalty0 (4):\penalty0 4396--4415, 2023.
\newblock Conference Name: IEEE Transactions on Pattern Analysis and Machine
  Intelligence.

\bibitem[Zhou(2012)]{zhou2012ensemble}
Zhi-Hua Zhou.
\newblock \emph{Ensemble methods: foundations and algorithms}.
\newblock CRC press, 2012.

\bibitem[Zhuang et~al.(2022)Zhuang, Gong, Yuan, Cui, Adam, Dvornek, sekhar
  tatikonda, s~Duncan, and Liu]{zhuang2022surrogate}
Juntang Zhuang, Boqing Gong, Liangzhe Yuan, Yin Cui, Hartwig Adam, Nicha~C
  Dvornek, sekhar tatikonda, James s Duncan, and Ting Liu.
\newblock Surrogate gap minimization improves sharpness-aware training.
\newblock In \emph{International Conference on Learning Representations}, 2022.

\end{thebibliography}


\begin{thebibliography}{59}
\providecommand{\natexlab}[1]{#1}
\providecommand{\url}[1]{\texttt{#1}}
\expandafter\ifx\csname urlstyle\endcsname\relax
  \providecommand{\doi}[1]{doi: #1}\else
  \providecommand{\doi}{doi: \begingroup \urlstyle{rm}\Url}\fi

\bibitem[Arjovsky et~al.(2019)Arjovsky, Bottou, Gulrajani, and
  Lopez-Paz]{arjovsky2019irm}
Martin Arjovsky, L{\'e}on Bottou, Ishaan Gulrajani, and David Lopez-Paz.
\newblock Invariant risk minimization.
\newblock \emph{arXiv preprint arXiv:1907.02893}, 2019.

\bibitem[Balaji et~al.(2018)Balaji, Sankaranarayanan, and
  Chellappa]{balaji2018metareg}
Yogesh Balaji, Swami Sankaranarayanan, and Rama Chellappa.
\newblock Metareg: Towards domain generalization using meta-regularization.
\newblock \emph{Advances in neural information processing systems}, 31, 2018.

\bibitem[Ballas and Diou(2024{\natexlab{a}})]{10233054}
Aristotelis Ballas and Christos Diou.
\newblock Towards domain generalization for ecg and eeg classification:
  Algorithms and benchmarks.
\newblock \emph{IEEE Transactions on Emerging Topics in Computational
  Intelligence}, 8\penalty0 (1):\penalty0 44--54, 2024{\natexlab{a}}.

\bibitem[Ballas and Diou(2024{\natexlab{b}})]{ballas2024multi}
Aristotelis Ballas and Christos Diou.
\newblock Multi-scale and multi-layer contrastive learning for domain
  generalization.
\newblock \emph{IEEE Transactions on Artificial Intelligence},
  2024{\natexlab{b}}.

\bibitem[Ben-David et~al.(2006)Ben-David, Blitzer, Crammer, and
  Pereira]{ben2006analysis}
Shai Ben-David, John Blitzer, Koby Crammer, and Fernando Pereira.
\newblock Analysis of representations for domain adaptation.
\newblock \emph{Advances in neural information processing systems}, 19, 2006.

\bibitem[Bengio et~al.(2013)Bengio, Courville, and Vincent]{6472238}
Yoshua Bengio, Aaron Courville, and Pascal Vincent.
\newblock Representation learning: A review and new perspectives.
\newblock \emph{IEEE Transactions on Pattern Analysis and Machine
  Intelligence}, 35\penalty0 (8):\penalty0 1798--1828, 2013.

\bibitem[Blanchard et~al.(2021)Blanchard, Deshmukh, Dogan, Lee, and
  Scott]{blanchard2021mtl_marginal_transfer_learning}
Gilles Blanchard, Aniket~Anand Deshmukh, Urun Dogan, Gyemin Lee, and Clayton
  Scott.
\newblock Domain generalization by marginal transfer learning.
\newblock \emph{Journal of Machine Learning Research}, 22\penalty0
  (2):\penalty0 1--55, 2021.

\bibitem[Borlino et~al.(2021)Borlino, D'Innocente, and
  Tommasi]{borlino2021rethinking}
Francesco~Cappio Borlino, Antonio D'Innocente, and Tatiana Tommasi.
\newblock Rethinking domain generalization baselines.
\newblock In \emph{2020 25th International Conference on Pattern Recognition
  (ICPR)}, pages 9227--9233. IEEE, 2021.

\bibitem[Carlucci et~al.(2019)Carlucci, D'Innocente, Bucci, Caputo, and
  Tommasi]{carlucci2019domain}
Fabio~M Carlucci, Antonio D'Innocente, Silvia Bucci, Barbara Caputo, and
  Tatiana Tommasi.
\newblock Domain generalization by solving jigsaw puzzles.
\newblock In \emph{Proceedings of the IEEE/CVF conference on computer vision
  and pattern recognition}, pages 2229--2238, 2019.

\bibitem[Cha et~al.(2021)Cha, Chun, Lee, Cho, Park, Lee, and Park]{cha2021swad}
Junbum Cha, Sanghyuk Chun, Kyungjae Lee, Han-Cheol Cho, Seunghyun Park, Yunsung
  Lee, and Sungrae Park.
\newblock Swad: Domain generalization by seeking flat minima.
\newblock \emph{Advances in Neural Information Processing Systems},
  34:\penalty0 22405--22418, 2021.

\bibitem[Chattopadhyay et~al.(2020)Chattopadhyay, Balaji, and
  Hoffman]{chattopadhyay2020learning}
Prithvijit Chattopadhyay, Yogesh Balaji, and Judy Hoffman.
\newblock Learning to balance specificity and invariance for in and out of
  domain generalization.
\newblock In \emph{Computer Vision--ECCV 2020: 16th European Conference,
  Glasgow, UK, August 23--28, 2020, Proceedings, Part IX 16}, pages 301--318.
  Springer, 2020.

\bibitem[Chen et~al.(2016)Chen, Duan, Houthooft, Schulman, Sutskever, and
  Abbeel]{chen2016infogan}
Xi Chen, Yan Duan, Rein Houthooft, John Schulman, Ilya Sutskever, and Pieter
  Abbeel.
\newblock Infogan: Interpretable representation learning by information
  maximizing generative adversarial nets.
\newblock \emph{Advances in neural information processing systems}, 29, 2016.

\bibitem[Choi et~al.(2023)Choi, Das, Choi, Yang, Park, and
  Yun]{choi2023progressive}
Seokeon Choi, Debasmit Das, Sungha Choi, Seunghan Yang, Hyunsin Park, and
  Sungrack Yun.
\newblock Progressive random convolutions for single domain generalization.
\newblock In \emph{Proceedings of the IEEE/CVF Conference on Computer Vision
  and Pattern Recognition}, pages 10312--10322, 2023.

\bibitem[Ding and Fu(2017)]{ding2017deep}
Zhengming Ding and Yun Fu.
\newblock Deep domain generalization with structured low-rank constraint.
\newblock \emph{IEEE Transactions on Image Processing}, 27\penalty0
  (1):\penalty0 304--313, 2017.

\bibitem[Dissanayake et~al.(2021)Dissanayake, Fernando, Denman, Ghaemmaghami,
  Sridharan, and Fookes]{9298838}
Theekshana Dissanayake, Tharindu Fernando, Simon Denman, Houman Ghaemmaghami,
  Sridha Sridharan, and Clinton Fookes.
\newblock Domain generalization in biosignal classification.
\newblock \emph{IEEE Transactions on Biomedical Engineering}, 68\penalty0
  (6):\penalty0 1978--1989, 2021.

\bibitem[Du et~al.(2021)Du, Wang, Feng, Pan, Qin, Xu, and Wang]{du2021adarnn}
Yuntao Du, Jindong Wang, Wenjie Feng, Sinno Pan, Tao Qin, Renjun Xu, and
  Chongjun Wang.
\newblock Adarnn: Adaptive learning and forecasting of time series.
\newblock In \emph{Proceedings of the 30th ACM international conference on
  information \& knowledge management}, pages 402--411, 2021.

\bibitem[Finn et~al.(2017)Finn, Abbeel, and Levine]{finn2017model}
Chelsea Finn, Pieter Abbeel, and Sergey Levine.
\newblock Model-agnostic meta-learning for fast adaptation of deep networks.
\newblock In \emph{International conference on machine learning}, pages
  1126--1135. PMLR, 2017.

\bibitem[Foret et~al.(2021)Foret, Kleiner, Mobahi, and
  Neyshabur]{foret2020sharpness}
Pierre Foret, Ariel Kleiner, Hossein Mobahi, and Behnam Neyshabur.
\newblock Sharpness-aware minimization for efficiently improving
  generalization.
\newblock In \emph{International Conference on Learning Representations}, 2021.

\bibitem[Gulrajani and Lopez-Paz(2021)]{gulrajani2020domainbed}
Ishaan Gulrajani and David Lopez-Paz.
\newblock In search of lost domain generalization.
\newblock In \emph{International Conference on Learning Representations}, 2021.

\bibitem[Huang and Belongie(2017)]{huang2017arbitrary}
Xun Huang and Serge Belongie.
\newblock Arbitrary style transfer in real-time with adaptive instance
  normalization.
\newblock In \emph{Proceedings of the IEEE international conference on computer
  vision}, pages 1501--1510, 2017.

\bibitem[Huang et~al.(2020)Huang, Wang, Xing, and Huang]{huang2020rsc}
Zeyi Huang, Haohan Wang, Eric~P Xing, and Dong Huang.
\newblock Self-challenging improves cross-domain generalization.
\newblock \emph{European Conference on Computer Vision}, 2020.

\bibitem[Hupkes et~al.(2023)Hupkes, Giulianelli, Dankers, Artetxe, Elazar,
  Pimentel, Christodoulopoulos, Lasri, Saphra, Sinclair,
  et~al.]{hupkes2023taxonomy}
Dieuwke Hupkes, Mario Giulianelli, Verna Dankers, Mikel Artetxe, Yanai Elazar,
  Tiago Pimentel, Christos Christodoulopoulos, Karim Lasri, Naomi Saphra,
  Arabella Sinclair, et~al.
\newblock A taxonomy and review of generalization research in nlp.
\newblock \emph{Nature Machine Intelligence}, 5\penalty0 (10):\penalty0
  1161--1174, 2023.

\bibitem[Ilse et~al.(2020)Ilse, Tomczak, Louizos, and Welling]{ilse2020diva}
Maximilian Ilse, Jakub~M Tomczak, Christos Louizos, and Max Welling.
\newblock Diva: Domain invariant variational autoencoders.
\newblock In \emph{Medical Imaging with Deep Learning}, pages 322--348. PMLR,
  2020.

\bibitem[Kim et~al.(2021)Kim, Yoo, Park, Kim, and Lee]{kim2021selfreg}
Daehee Kim, Youngjun Yoo, Seunghyun Park, Jinkyu Kim, and Jaekoo Lee.
\newblock Selfreg: Self-supervised contrastive regularization for domain
  generalization.
\newblock In \emph{International Conference on Computer Vision}, 2021.

\bibitem[Krueger et~al.(2021{\natexlab{a}})Krueger, Caballero, Jacobsen, Zhang,
  Binas, Zhang, Le~Priol, and Courville]{krueger2020vrex}
David Krueger, Ethan Caballero, Joern-Henrik Jacobsen, Amy Zhang, Jonathan
  Binas, Dinghuai Zhang, Remi Le~Priol, and Aaron Courville.
\newblock Out-of-distribution generalization via risk extrapolation (rex).
\newblock In \emph{International conference on machine learning}, pages
  5815--5826. PMLR, 2021{\natexlab{a}}.

\bibitem[Krueger et~al.(2021{\natexlab{b}})Krueger, Caballero, Jacobsen, Zhang,
  Binas, Zhang, Le~Priol, and Courville]{krueger2021out}
David Krueger, Ethan Caballero, Joern-Henrik Jacobsen, Amy Zhang, Jonathan
  Binas, Dinghuai Zhang, Remi Le~Priol, and Aaron Courville.
\newblock Out-of-distribution generalization via risk extrapolation (rex).
\newblock In \emph{International conference on machine learning}, pages
  5815--5826. PMLR, 2021{\natexlab{b}}.

\bibitem[Laskin et~al.(2020)Laskin, Srinivas, and Abbeel]{laskin2020curl}
Michael Laskin, Aravind Srinivas, and Pieter Abbeel.
\newblock Curl: Contrastive unsupervised representations for reinforcement
  learning.
\newblock In \emph{International conference on machine learning}, pages
  5639--5650. PMLR, 2020.

\bibitem[Lee et~al.(2019)Lee, Lee, Shin, and Lee]{lee2019network}
Kimin Lee, Kibok Lee, Jinwoo Shin, and Honglak Lee.
\newblock Network randomization: A simple technique for generalization in deep
  reinforcement learning.
\newblock \emph{arXiv preprint arXiv:1910.05396}, 2019.

\bibitem[Li et~al.(2018{\natexlab{a}})Li, Yang, Song, and
  Hospedales]{li2018learning}
Da Li, Yongxin Yang, Yi-Zhe Song, and Timothy Hospedales.
\newblock Learning to generalize: Meta-learning for domain generalization.
\newblock In \emph{Proceedings of the AAAI conference on artificial
  intelligence}, 2018{\natexlab{a}}.

\bibitem[Li et~al.(2018{\natexlab{b}})Li, Pan, Wang, and Kot]{li2018mmd}
Haoliang Li, Sinno~Jialin Pan, Shiqi Wang, and Alex~C Kot.
\newblock Domain generalization with adversarial feature learning.
\newblock In \emph{Computer Vision and Pattern Recognition},
  2018{\natexlab{b}}.

\bibitem[Li et~al.(2018{\natexlab{c}})Li, Tian, Gong, Liu, Liu, Zhang, and
  Tao]{li2018deep}
Ya Li, Xinmei Tian, Mingming Gong, Yajing Liu, Tongliang Liu, Kun Zhang, and
  Dacheng Tao.
\newblock Deep domain generalization via conditional invariant adversarial
  networks.
\newblock In \emph{Proceedings of the European conference on computer vision
  (ECCV)}, pages 624--639, 2018{\natexlab{c}}.

\bibitem[Mahajan et~al.(2021)Mahajan, Tople, and Sharma]{mahajan2021domain}
Divyat Mahajan, Shruti Tople, and Amit Sharma.
\newblock Domain generalization using causal matching.
\newblock In \emph{International conference on machine learning}, pages
  7313--7324. PMLR, 2021.

\bibitem[Matsuura and Harada(2020)]{matsuura2020domain}
Toshihiko Matsuura and Tatsuya Harada.
\newblock Domain generalization using a mixture of multiple latent domains.
\newblock In \emph{Proceedings of the AAAI conference on artificial
  intelligence}, pages 11749--11756, 2020.

\bibitem[Nam et~al.(2021)Nam, Lee, Park, Yoon, and Yoo]{nam2019sagnet}
Hyeonseob Nam, HyunJae Lee, Jongchan Park, Wonjun Yoon, and Donggeun Yoo.
\newblock Reducing domain gap by reducing style bias.
\newblock In \emph{Computer Vision and Pattern Recognition}, 2021.

\bibitem[Ot{\'a}lora et~al.(2019)Ot{\'a}lora, Atzori, Andrearczyk, Khan, and
  M{\"u}ller]{otalora2019staining}
Sebastian Ot{\'a}lora, Manfredo Atzori, Vincent Andrearczyk, Amjad Khan, and
  Henning M{\"u}ller.
\newblock Staining invariant features for improving generalization of deep
  convolutional neural networks in computational pathology.
\newblock \emph{Frontiers in bioengineering and biotechnology}, 7:\penalty0
  198, 2019.

\bibitem[Piratla et~al.(2020)Piratla, Netrapalli, and
  Sarawagi]{piratla2020efficient}
Vihari Piratla, Praneeth Netrapalli, and Sunita Sarawagi.
\newblock Efficient domain generalization via common-specific low-rank
  decomposition.
\newblock In \emph{International conference on machine learning}, pages
  7728--7738. PMLR, 2020.

\bibitem[Qiao et~al.(2020)Qiao, Zhao, and Peng]{qiao2020learning}
Fengchun Qiao, Long Zhao, and Xi Peng.
\newblock Learning to learn single domain generalization.
\newblock In \emph{Proceedings of the IEEE/CVF conference on computer vision
  and pattern recognition}, pages 12556--12565, 2020.

\bibitem[Sagawa* et~al.(2020)Sagawa*, Koh*, Hashimoto, and
  Liang]{Sagawa2020GroupDRO}
Shiori Sagawa*, Pang~Wei Koh*, Tatsunori~B. Hashimoto, and Percy Liang.
\newblock Distributionally robust neural networks.
\newblock In \emph{International Conference on Learning Representations}, 2020.

\bibitem[Shankar et~al.(2018)Shankar, Piratla, Chakrabarti, Chaudhuri, Jyothi,
  and Sarawagi]{shankar2018generalizing}
Shiv Shankar, Vihari Piratla, Soumen Chakrabarti, Siddhartha Chaudhuri, Preethi
  Jyothi, and Sunita Sarawagi.
\newblock Generalizing across domains via cross-gradient training.
\newblock \emph{arXiv preprint arXiv:1804.10745}, 2018.

\bibitem[Shi et~al.(2022)Shi, Seely, Torr, N, Hannun, Usunier, and
  Synnaeve]{shi2021gradient}
Yuge Shi, Jeffrey Seely, Philip Torr, Siddharth N, Awni Hannun, Nicolas
  Usunier, and Gabriel Synnaeve.
\newblock Gradient matching for domain generalization.
\newblock In \emph{International Conference on Learning Representations}, 2022.

\bibitem[Sun and Saenko(2016)]{sun2016coral}
Baochen Sun and Kate Saenko.
\newblock Deep coral: Correlation alignment for deep domain adaptation.
\newblock In \emph{European Conference on Computer Vision}, 2016.

\bibitem[Tobin et~al.(2017)Tobin, Fong, Ray, Schneider, Zaremba, and
  Abbeel]{tobin2017domain}
Josh Tobin, Rachel Fong, Alex Ray, Jonas Schneider, Wojciech Zaremba, and
  Pieter Abbeel.
\newblock Domain randomization for transferring deep neural networks from
  simulation to the real world.
\newblock In \emph{2017 IEEE/RSJ international conference on intelligent robots
  and systems (IROS)}, pages 23--30. IEEE, 2017.

\bibitem[Vapnik(1998)]{vapnik1998statistical}
V Vapnik.
\newblock Statistical learning theory.
\newblock \emph{NY: Wiley}, 1998.

\bibitem[Volpi and Murino(2019)]{volpi2019addressing}
Riccardo Volpi and Vittorio Murino.
\newblock Addressing model vulnerability to distributional shifts over image
  transformation sets.
\newblock In \emph{Proceedings of the IEEE/CVF International Conference on
  Computer Vision}, pages 7980--7989, 2019.

\bibitem[Volpi et~al.(2018)Volpi, Namkoong, Sener, Duchi, Murino, and
  Savarese]{volpi2018generalizing}
Riccardo Volpi, Hongseok Namkoong, Ozan Sener, John~C Duchi, Vittorio Murino,
  and Silvio Savarese.
\newblock Generalizing to unseen domains via adversarial data augmentation.
\newblock \emph{Advances in neural information processing systems}, 31, 2018.

\bibitem[Wang et~al.(2022)Wang, Lan, Liu, Ouyang, Qin, Lu, Chen, Zeng, and
  Philip]{wang2022generalizing}
Jindong Wang, Cuiling Lan, Chang Liu, Yidong Ouyang, Tao Qin, Wang Lu, Yiqiang
  Chen, Wenjun Zeng, and S~Yu Philip.
\newblock Generalizing to unseen domains: A survey on domain generalization.
\newblock \emph{IEEE transactions on knowledge and data engineering},
  35\penalty0 (8):\penalty0 8052--8072, 2022.

\bibitem[Wang et~al.(2024)Wang, Ren, Shen, Huang, and Zhu]{wang2024multi}
Jun Wang, He Ren, Changqing Shen, Weiguo Huang, and Zhongkui Zhu.
\newblock Multi-scale style generative and adversarial contrastive networks for
  single domain generalization fault diagnosis.
\newblock \emph{Reliability Engineering \& System Safety}, 243:\penalty0
  109879, 2024.

\bibitem[Wang et~al.(2023)Wang, Zhang, Lei, and Zhang]{wang2023sharpness}
Pengfei Wang, Zhaoxiang Zhang, Zhen Lei, and Lei Zhang.
\newblock Sharpness-aware gradient matching for domain generalization.
\newblock In \emph{Proceedings of the IEEE/CVF Conference on Computer Vision
  and Pattern Recognition}, pages 3769--3778, 2023.

\bibitem[Wang et~al.(2020)Wang, Yu, Li, Yang, Fu, and Heng]{wang2020dofe}
Shujun Wang, Lequan Yu, Kang Li, Xin Yang, Chi-Wing Fu, and Pheng-Ann Heng.
\newblock Dofe: Domain-oriented feature embedding for generalizable fundus
  image segmentation on unseen datasets.
\newblock \emph{IEEE Transactions on Medical Imaging}, 39\penalty0
  (12):\penalty0 4237--4248, 2020.

\bibitem[Xu et~al.(2020)Xu, Zhang, Ni, Li, Wang, Tian, and
  Zhang]{xu2020interdomain_mixup_aaai}
Minghao Xu, Jian Zhang, Bingbing Ni, Teng Li, Chengjie Wang, Qi Tian, and
  Wenjun Zhang.
\newblock Adversarial domain adaptation with domain mixup.
\newblock In \emph{AAAI Conference on Artificial Intelligence}, 2020.

\bibitem[Yosinski et~al.(2014)Yosinski, Clune, Bengio, and
  Lipson]{yosinski2014transferable}
Jason Yosinski, Jeff Clune, Yoshua Bengio, and Hod Lipson.
\newblock How transferable are features in deep neural networks?
\newblock \emph{Advances in neural information processing systems}, 27, 2014.

\bibitem[Yue et~al.(2019)Yue, Zhang, Zhao, Sangiovanni-Vincentelli, Keutzer,
  and Gong]{yue2019domain}
Xiangyu Yue, Yang Zhang, Sicheng Zhao, Alberto Sangiovanni-Vincentelli, Kurt
  Keutzer, and Boqing Gong.
\newblock Domain randomization and pyramid consistency: Simulation-to-real
  generalization without accessing target domain data.
\newblock In \emph{Proceedings of the IEEE/CVF international conference on
  computer vision}, pages 2100--2110, 2019.

\bibitem[Zhang et~al.(2022)Zhang, Zhang, Liu, Weller, Sch\"olkopf, and
  Xing]{Zhang_2022_CVPR}
Hanlin Zhang, Yi-Fan Zhang, Weiyang Liu, Adrian Weller, Bernhard Sch\"olkopf,
  and Eric~P. Xing.
\newblock Towards principled disentanglement for domain generalization.
\newblock In \emph{Proceedings of the IEEE/CVF Conference on Computer Vision
  and Pattern Recognition (CVPR)}, pages 8024--8034, 2022.

\bibitem[Zhang et~al.(2020{\natexlab{a}})Zhang, Wang, Yang, Sanford, Harmon,
  Turkbey, Wood, Roth, Myronenko, Xu, et~al.]{zhang2020generalizing}
Ling Zhang, Xiaosong Wang, Dong Yang, Thomas Sanford, Stephanie Harmon, Baris
  Turkbey, Bradford~J Wood, Holger Roth, Andriy Myronenko, Daguang Xu, et~al.
\newblock Generalizing deep learning for medical image segmentation to unseen
  domains via deep stacked transformation.
\newblock \emph{IEEE transactions on medical imaging}, 39\penalty0
  (7):\penalty0 2531--2540, 2020{\natexlab{a}}.

\bibitem[Zhang et~al.(2020{\natexlab{b}})Zhang, Marklund, Gupta, Levine, and
  Finn]{zhang2020arm}
Marvin Zhang, Henrik Marklund, Abhishek Gupta, Sergey Levine, and Chelsea Finn.
\newblock Adaptive risk minimization: A meta-learning approach for tackling
  group shift.
\newblock \emph{arXiv preprint arXiv:2007.02931}, 2020{\natexlab{b}}.

\bibitem[Zheng et~al.(2021)Zheng, Yang, Yin, Li, Wang, and Xu]{9174912}
Huailiang Zheng, Yuantao Yang, Jiancheng Yin, Yuqing Li, Rixin Wang, and
  Minqiang Xu.
\newblock Deep domain generalization combining a priori diagnosis knowledge
  toward cross-domain fault diagnosis of rolling bearing.
\newblock \emph{IEEE Transactions on Instrumentation and Measurement},
  70:\penalty0 1--11, 2021.

\bibitem[Zhou et~al.(2021{\natexlab{a}})Zhou, Yang, Qiao, and
  Xiang]{zhou2021domain}
Kaiyang Zhou, Yongxin Yang, Yu Qiao, and Tao Xiang.
\newblock Domain adaptive ensemble learning.
\newblock \emph{IEEE Transactions on Image Processing}, 30:\penalty0
  8008--8018, 2021{\natexlab{a}}.

\bibitem[Zhou et~al.(2021{\natexlab{b}})Zhou, Yang, Qiao, and
  Xiang]{zhou2021mixstyle}
Kaiyang Zhou, Yongxin Yang, Yu Qiao, and Tao Xiang.
\newblock Domain generalization with mixstyle.
\newblock In \emph{International Conference on Learning Representations},
  2021{\natexlab{b}}.

\bibitem[Zhuang et~al.(2022)Zhuang, Gong, Yuan, Cui, Adam, Dvornek, sekhar
  tatikonda, s~Duncan, and Liu]{zhuang2022surrogate}
Juntang Zhuang, Boqing Gong, Liangzhe Yuan, Yin Cui, Hartwig Adam, Nicha~C
  Dvornek, sekhar tatikonda, James s Duncan, and Ting Liu.
\newblock Surrogate gap minimization improves sharpness-aware training.
\newblock In \emph{International Conference on Learning Representations}, 2022.

\end{thebibliography}



\end{document}